\documentclass[10pt,twocolumn,letterpaper]{article}

\usepackage{cvpr} 

\usepackage{graphicx}
\usepackage{amsmath}
\usepackage{amssymb}
\usepackage{booktabs}

\usepackage{url}
\usepackage{multicol}
\usepackage{wrapfig}
\usepackage{lipsum}
\usepackage{algorithm}
\usepackage{algcompatible}
\usepackage{xcolor}
\usepackage{multirow}

\usepackage{comment}

\newtheorem{assumption}{Assumption}
\newtheorem{remark}{Remark}
\newtheorem{proposition}{Proposition}

\usepackage{colortbl}
\usepackage{xcolor}
\definecolor{LightCyan}{rgb}{0.88,1,1}

\usepackage[pagebackref,breaklinks,colorlinks]{hyperref}

\usepackage[capitalize]{cleveref}
\crefname{section}{Sec.}{Secs.}
\Crefname{section}{Section}{Sections}
\Crefname{table}{Table}{Tables}
\crefname{table}{Tab.}{Tabs.}

\def\cvprPaperID{11877} 
\def\confName{CVPR}
\def\confYear{2023}

\begin{document}

\title{Lookahead Diffusion Probabilistic Models for Refining Mean Estimation}

\author{Guoqiang Zhang\\
University of Technology Sydney\\
{\tt\small guoqiang.zhang@uts.edu.au }
\and
Kenta Niwa\\
NTT Communication Science Laboratories\\
{\tt\small kenta.niwa.bk@hco.ntt.co.jp}
\and 
W. Bastiaan Kleijn \\
Victoria University of Wellington \\
{\tt\small bastiaan.kleijn@vuw.ac.nz}
}

\maketitle

\begin{abstract}
We propose lookahead diffusion probabilistic models (LA-DPMs) to exploit the correlation in the outputs of the deep neural networks (DNNs) over subsequent timesteps in diffusion probabilistic models (DPMs) to refine the mean estimation of the conditional Gaussian distributions in the backward process. A typical DPM first obtains an estimate of the original data sample $\boldsymbol{x}$ by feeding the most recent state $\boldsymbol{z}_i$ and index $i$ into the DNN model and then computes the mean vector of the conditional Gaussian distribution for $\boldsymbol{z}_{i-1}$. We propose to calculate a more accurate estimate for $\boldsymbol{x}$ by performing extrapolation on the two estimates of $\boldsymbol{x}$ that are obtained by feeding $(\boldsymbol{z}_{i+1},i+1)$ and $(\boldsymbol{z}_{i},i)$ into the DNN model. The extrapolation can be  easily integrated into the backward process of existing DPMs by introducing an additional connection over two consecutive timesteps, and fine-tuning is not required. Extensive experiments showed that plugging in the additional connection into DDPM, DDIM, DEIS, S-PNDM, and high-order DPM-Solvers leads to a significant performance gain in terms of Fréchet inception distance (FID) score. Our implementation is available at \url{https://github.com/guoqiang-zhang-x/LA-DPM}.

\end{abstract}


\vspace{-3.5mm}
\section{Introduction}
\vspace{-1mm}
As one type of generative model, diffusion probabilistic models (DPMs) have made significant progress in recent years. The pioneering work \cite{Dickstein15DPM} applied non-equilibrium statistical physics to estimating probabilistic data distributions. In doing so, a Markov forward diffusion process is constructed by systematically inserting additive noise in the data until essentially only noise remains. The data distribution is then gradually restored by a reverse diffusion process starting from a simple parametric distribution. The main advantage of DPMs over classic tractable models (e.g., HMMs, GMMs, see \cite{Bishop06}) is that they can accurately model both the high and low likelihood regions of the data distribution via the progressive estimation of noise-perturbed data distributions. In comparison to generative adversarial networks (GANs) \cite{Goodfellow14GAN, Arjovsky17WGAN, Gulrajani17WGANGP}, DPMs exhibit more stable training dynamics by avoiding adversarial learning. 

The work \cite{Ho20DDPM} focuses on a particular type of DPM, namely a denoising diffusion  probabilistic model (DDPM), and shows that after a sufficient number of timesteps (or equivalently iterations) in the backward process, DDPM can achieve state-of-the-art performance in image generation tasks by the proper design of a weighted variational bound (VB). In addition, by inspection of the weighted VB, it is found that the method \emph{score matching with Langevin dynamics} (SMLD) \cite{Song19, Song21DPM} can also be viewed as a DPM. The recent work \cite{Song21SDE_gen} interprets DDPM and SMLD as search of approximate solutions to stochastic differential equations. See also \cite{Nichol21DDPM} and \cite{Dhariwal21DPM} for improved DPMs that lead to better log-likelihood scores and sampling qualities, respectively.  

One inconvenience of a standard DPM is that the associated deep neural network (DNN) needs to run for a sufficient number of timesteps to achieve high sampling quality while the generative model of a GAN only needs to run once. This has led to an increasing research focus on reducing the number of reverse timesteps in DPMs while retaining a satisfactory sampling quality (see \cite{Yang21DPM} for a detailed overview). Song et al. proposed the so-called denoising diffusion implicit model (DDIM) \cite{Song21DDIM} as an extension of DDPM from a non-Markov forward process point of view. The work \cite{Kingma21DDPM} proposed to learn a denoising schedule in the reverse process by explicitly modeling the signal-to-noise ratio in the image generation task. \cite{Chen20WaveGrad} and
\cite{Lam22BDDM} considered effective audio generation by proposing different noise scheduling schemes in DPMs. 
Differently from the above methods, the recent works \cite{Bao22DPM} and \cite{Bao22DPM_cov} proposed to estimate the optimal variance of the backward conditional Gaussian distribution to improve sampling qualities for both small and large numbers of timesteps.


Another approach for improving the sampling quality of DPMs with a limited computational budget is to exploit high-order methods for solving the backward ordinary differential equations (ODEs) (see \cite{Song21SDE_gen}). The authors of \cite{Liu22PNDM} proposed pseudo numerical methods for diffusion models (PNDM), of which high-order polynomials of the estimated Gaussian noises $\{\hat{\boldsymbol{\epsilon}}_{\boldsymbol{\theta}}(\boldsymbol{z}_{i+j},i+j) | r\geq j\geq 0\}$  are introduced to better estimate the latent variable $\boldsymbol{z}_{i-1}$ at iteration $i$, where $ \hat{\boldsymbol{\epsilon}}_{\boldsymbol{\theta}}$ represents a pre-trained neural network model for predicting the Gaussian noises. The work \cite{Zhang22DEIS} further extends \cite{Liu22PNDM} by refining the coefficients of the high-order polynomials of the estimated Gaussian noises, and proposes the diffusion exponential integrator sampler (DEIS). Recently, the authors of \cite{Lu22DPM_Solver} considered solving the ODEs of a diffusion model differently from \cite{Zhang22DEIS}. In particular, a high-order Taylor expansion of the estimated Gaussian noises was employed to better approximate the continuous solutions of the ODEs, where the developed sampling methods are referred to as DPM-Solvers.


We note that the computation of $\boldsymbol{z}_{i-1}$ at timestep $i$ in the backward process of existing DPMs (including the high-order ODE solvers) can always be reformulated in terms of an estimate $\hat{\boldsymbol{x}}$ for the original data sample $\boldsymbol{x}$ in combination with other terms. In principle, as the timestep $i$ decreases, the estimate  $\hat{\boldsymbol{x}}$ would become increasingly accurate. 
In this paper, we aim to improve the estimation accuracy of $\boldsymbol{x}$ at each timestep $i$ in computation of the mean vector for the latent variable $\boldsymbol{z}_{i-1}$. To do so, we propose to make an extrapolation from the two most recent estimates of $\boldsymbol{x}$ obtained at timestep $i$ and $i+1$. The extrapolation allows the backward process to look ahead towards a noisy direction targeting $\boldsymbol{x}$, thus improving the estimation accuracy. The extrapolation can be realized by simply introducing additional connections between two consecutive timesteps, which can be easily plugged into existing DPMs with negligible computational overhead. We refer to the improved diffusion models as Lookahead-DPMs (LA-DPMs).           

\vspace{-0.5mm}
We conducted an extensive evaluation by plugging in the additional connection into the backward process of DDPM, DDIM, DEIS, S-PNDM, and DPM-Solver. Interestingly, it is found that the performance gain of LA-DPMs is more significant for a small number of timesteps. This makes it attractive for practical applications as it is computationally preferable to run the backward process in a limited number of timesteps.


\vspace{-1.0mm}
\section{Background of Markov Diffusion Models}
\vspace{-0.5mm}

We revisit the standard Markov DPMs being studied in \cite{Kingma21DDPM}. In the following, we first briefly review the forward diffusion process. We then investigate its backward process. 
The notation in this paper is in line with that of \cite{Kingma21DDPM}. 

\subsection{Forward diffusion process}
\label{subsec:forward}
Suppose we have a set of observations of $\boldsymbol{x}$ that are drawn from a data distribution $q(\boldsymbol{x})$. A forward diffusion process can be defined as a sequence of increasingly noisy versions $\boldsymbol{z}_t$, $t\in[0,1]$, of $\boldsymbol{x}$, where $\boldsymbol{z}_{t=1}$ indicates the noisiest version (we will discretize $t$ later on). For a Gaussian-driven process,  the latent variable $\boldsymbol{z}_t$ can be represented in terms of $\boldsymbol{x}$ being contaminated by a Gaussian noise as  
\begin{align}
    \boldsymbol{z}_t =\alpha_t\boldsymbol{x} + \sigma_t\boldsymbol{\epsilon}_t, \label{equ:ForwardGau}
\end{align}
where $\boldsymbol{\epsilon}_t\sim \mathcal{N}(0,\boldsymbol{I})$, and $\alpha_t$ and $\sigma_t$ are strictly positive scalar-valued functions of $t$, and $\boldsymbol{I}$ is the identity matrix. 
To formalise the notion of $\boldsymbol{z}_t$ being increasingly noisy, $\boldsymbol{z}_t$ can be alternatively represented in terms of $\boldsymbol{z}_s$, $s<t$, as
\begin{align}
\boldsymbol{z}_t &= \alpha_{t|s}\boldsymbol{z}_s +\sigma_{t|s}\boldsymbol{\epsilon}_{t|s},
\label{equ:ForwardGauAdd0}
\end{align}
where $\boldsymbol{\epsilon}_{t|s}$ is the additional Gaussian noise being added to a scaled version of $\boldsymbol{z}_s$, and $(\alpha_{t|s},\sigma_{t|s}^2)$ are given by
\begin{align}
\alpha_{t|s}=\alpha_t/\alpha_s \;\textrm{ and }\; \sigma_{t|s}^2=\sigma_t^2-\alpha_{t|s}^2 \sigma_s^2, \label{equ:ForwardGauAdd} 
\end{align}
where the conditional variance $\sigma_{t|s}^2$ is assume to be positive, i.e., $\sigma_{t|s}^2>0$.
One major advantage of the above formulation is that it includes both the variance-preserving process with $\alpha_t=\sqrt{1-\sigma_t^2}$ \cite{Dickstein15DPM,Ho20DDPM} and variance-exploding process with $\alpha_t=1$ \cite{Song19,Song21SDE_gen}.

It is immediate that the process (\ref{equ:ForwardGau})-(\ref{equ:ForwardGauAdd}) is Markov. That is, the conditional distribution $q(\boldsymbol{z}_u|\boldsymbol{z}_t,\boldsymbol{z}_s)=q(\boldsymbol{z}_u|\boldsymbol{z}_t)=\mathcal{N}(\alpha_{u|t}\boldsymbol{z}_t, \sigma_{u|t}^2\boldsymbol{I})$, where $0\leq s<t<u\leq 1$. Consequently, it can be shown that $q(\boldsymbol{z}_s | \boldsymbol{z}_t, \boldsymbol{x})$, $s<t$, is Normal distributed by using Bayes rule (see Appendix~A of \cite{Kingma21DDPM}), 
\begin{equation}
\hspace{-3mm}q(\boldsymbol{z}_s | \boldsymbol{z}_t, \boldsymbol{x}) = \mathcal{N}\left(\frac{\sigma_s^2}{\sigma_t^2}\alpha_{t|s}\boldsymbol{z}_t+\frac{\sigma_{t|s}^2}{\sigma_t^2}\alpha_s\boldsymbol{x}, \frac{\sigma_s^2\sigma_{t|s}^2}{\sigma_t^2}\boldsymbol{I}\right). \label{equ:midPred}
\end{equation}
As will be discussed later on, the backward process heavily relies on the relation between $(\boldsymbol{z}_t, \boldsymbol{x})$ and $\boldsymbol{z_s}$ in the formulation (\ref{equ:midPred}). 

As one example, the above process includes the forward process of a DDPM as a special case. One can simply discretize $t\in[0,1]$ into $N$ uniform timesteps, i.e., $t_i=i/N$, and let $\{\alpha_{t_i}=\sqrt{1-\sigma_{t_i}^2}| N\geq i\geq 0\}$ be a strictly decreasing sequence.

\subsection{Backward diffusion process}
In general, a backward process is designed to reverse the forward process introduced earlier for the purpose of approximating the data distribution $q(\boldsymbol{x})$. Without loss of generality, we denote a discrete backward process as  
\begin{align}
p(\boldsymbol{x},\boldsymbol{z}_{0:N}) = p(\boldsymbol{z}_{N})\prod_{i=1}^N p(\boldsymbol{z}_{i-1}|\boldsymbol{z}_{i:N}) p(\boldsymbol{x}|\boldsymbol{z}_{0:N}),
\label{equ:backward}
\end{align}
where the support region $[0,1]$ for $t$ is discretized into $N$ uniform timesteps, i.e., $t_i=i/N$, and $t_i$ is replaced by $i$ to simplify notation. The objective is to find a specific form of the backward process such that its marginal distribution with regard to $\boldsymbol{x}$ approaches $q(\boldsymbol{x})$:
\begin{align}
   q(\boldsymbol{x})\approx \int p(\boldsymbol{x},\boldsymbol{z}_{0:N})d\boldsymbol{z}_0\ldots d\boldsymbol{z}_N.  \label{equ:obj}
\end{align}

To facilitate computation, DDPM makes the following approximation to the backward process %
of (\ref{equ:midPred}): 
\begin{align}
 &p(\boldsymbol{z}_{i-1}|\boldsymbol{z}_{i:N})  \nonumber \\ 
 &= p(\boldsymbol{z}_{i-1}|\boldsymbol{z}_{i}) 
 \label{equ:Gausian0} \\
 &\approx q(\boldsymbol{z}_{i-1}|  \boldsymbol{z}_i,\boldsymbol{x} =\hat{\boldsymbol{x}}(\boldsymbol{z}_i,i))
 \notag \\
 &= \hspace{-0.6mm}\mathcal{N}\hspace{-0.6mm}\left(\hspace{-0.6mm}\underbrace{\frac{\sigma_{i-1}^2}{\sigma_i^2}\alpha_{i|i-1}\boldsymbol{z}_i\hspace{-0.6mm}+\hspace{-0.6mm}\frac{\sigma_{i|i-1}^2}{\sigma_i^2}\alpha_{i-1}\hat{\boldsymbol{x}}(\boldsymbol{z}_i, i)}_{\boldsymbol{\mu}(\boldsymbol{z}_{i-1}|\boldsymbol{z}_i,i)},\hspace{-0.6mm} \underbrace{\frac{\sigma_{i-1}^2\sigma_{i|i-1}^2}{\sigma_i^2}}_{\varphi_i}\hspace{-0.6mm}\boldsymbol{I}\hspace{-0.6mm}\right)\hspace{-0.8mm}, \label{equ:Gausian}
\end{align}
where $\alpha_{i|i-1}$ and $\sigma^2_{i|i-1}$ follow from (\ref{equ:ForwardGauAdd}) with $(t,s)=(i/N,(i-1)/N)$, and
$\hat{\boldsymbol{x}}(\boldsymbol{z}_i,i)$ denotes the predicted sample for $\boldsymbol{x}$ by using $\boldsymbol{z}_i$ and timestep $i$. The marginal distribution of $\boldsymbol{z}_N$ is approximated to be a spherical Gaussian, i.e.,  $p(\boldsymbol{z}_N)\approx p(0,\beta\boldsymbol{I})$, where $\beta=1$ for variance-preserving DPMs. A nice property of (\ref{equ:Gausian}) is that the conditional distribution is Gaussian. As a result, the computation in the backward process only needs to focus on a sequence of means $\boldsymbol{\mu}(\boldsymbol{z}_{i-1}|\boldsymbol{z}_i,i)$ and variances $\varphi_i$ from $i=N$ to $i=0$. 

Next, we briefly discuss the computation for $\hat{\boldsymbol{x}}(\boldsymbol{z}_i,i)$ in \cite{Ho20DDPM}, which is followed by recent, more advanced, DPM models such as DDIM \cite{Song21DDIM} and DPM-Solver \cite{Lu22DPM_Solver}.  In \cite{Ho20DDPM}, a DNN model $ \hat{\boldsymbol{\epsilon}}_{\boldsymbol{\theta}}$ is designed to make a direct prediction of the added Gaussian noise $\boldsymbol{\epsilon}_t$ to $\boldsymbol{x}$ in a latent variable $\boldsymbol{z}_t$ of (\ref{equ:ForwardGau}). In particular, the model is trained to minimize a summation of expected squared errors: 
\begin{align}
\min_{\boldsymbol{\theta}} \sum_{i=1}^N \mathbf{E}_{\boldsymbol{x}, \boldsymbol{\epsilon}_i}\left[  \|\hat{\boldsymbol{\epsilon}}_{\boldsymbol{\theta}}({\alpha}_i\boldsymbol{x}+\sqrt{1-{\alpha}_i^2}\boldsymbol{\epsilon}_i, i) -\boldsymbol{\epsilon}_i\|^2\right].
\label{equ:thetaMin}
\end{align}
As $\boldsymbol{\epsilon}$ and $\boldsymbol{z}$ share the same dimensionality, the architecture of the model $ \hat{\boldsymbol{\epsilon}}_{\boldsymbol{\theta}}$ is often selected to be a variant of UNet \cite{Ronneberger15Unet}. 
In the sampling process, an approximation of $\boldsymbol{x}$ can  be easily obtained in terms of $\hat{\boldsymbol{\epsilon}}_{\boldsymbol{\theta}}(\boldsymbol{z}_i, i)$ by following (\ref{equ:ForwardGau}) under the condition $\sigma_i= \sqrt{1-\alpha_i^2}$: 
\begin{align}
\hat{\boldsymbol{x}}(\boldsymbol{z}_i,i)&=\hat{\boldsymbol{x}}(\boldsymbol{z}_i,  \hat{\boldsymbol{\epsilon}}_{\boldsymbol{\theta}}(\boldsymbol{z}_i, i))\nonumber \\
&= \boldsymbol{z}_i/\alpha_i-\sqrt{1-\alpha_i^2}\hat{\boldsymbol{\epsilon}}_{\boldsymbol{\theta}}(\boldsymbol{z}_i, i)/\alpha_i. \label{equ:x_epsilon}
\end{align}
The expression (\ref{equ:x_epsilon}) for $\boldsymbol{x}$ can then be plugged into $\boldsymbol{\mu}(\boldsymbol{z}_{i-1}|\boldsymbol{z}_i, i)$ in (\ref{equ:Gausian}).


In practice, different approximations have been made to the variance $\varphi_i$ of the conditional Gaussian distribution  in  (\ref{equ:Gausian}). For instance, it has been found in \cite{Ho20DDPM} that two different setups for $\varphi_i$ lead to similar sampling performance. As mentioned earlier, the two recent works  \cite{Bao22DPM} and \cite{Bao22DPM_cov} propose to train DNNs to optimally estimate the time-dependent variance in (\ref{equ:Gausian}) under different conditions, which is found to produce considerably high sampling quality.  

\section{Basic Lookahead Diffusion Models} 
In this section, we first consider the correlations carried in $\{\hat{\boldsymbol{x}}(\boldsymbol{z}_j,\hat{\boldsymbol{\epsilon}}_{\boldsymbol{\theta}}(\boldsymbol{z}_j,j))| N\geq j\geq 0\}$ over consecutive timesteps. Then, we propose to refine the estimate for $\boldsymbol{x}$ by performing extrapolation in the backward process of DDPM, DDIM, and DPM-Solvers, respectively. We refer to the improved generative models as LA-DPMs. 
Finally, we conduct an analysis to study the strengths of the extrapolations.  

\subsection{Inspection of the estimates for $\boldsymbol{x}$ }

\begin{figure}[t!]
\centering
\includegraphics[width=70mm]{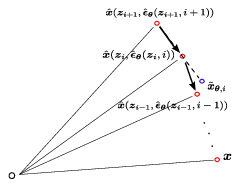}
\vspace*{-0.1cm}
\caption{\small{Illustration of the extrapolation operation for refining the mean estimation in the backward process of DDPM. At timestep $i$, the estimate $\tilde{\boldsymbol{x}}_{\boldsymbol{\theta},i}$ is computed by extrapolating from the two traditional estimates $\hat{\boldsymbol{x}}( \boldsymbol{z}_i, \hat{\boldsymbol{\epsilon}}_{\boldsymbol{\theta}}(\boldsymbol{z}_i,i))$ and $\hat{\boldsymbol{x}}( \boldsymbol{z}_{i+1}, \hat{\boldsymbol{\epsilon}}_{\boldsymbol{\theta}}(\boldsymbol{z}_{i+1},i+1))$. $\tilde{\boldsymbol{x}}_{\boldsymbol{\theta},i}$ is taken to replace $\boldsymbol{x}$ in the conditional Gaussian distribution $q(\boldsymbol{z}_{i-1}| \boldsymbol{z}_i, \boldsymbol{x})$.    } }
\label{fig:extraP}
\vspace*{-0.2cm}
\end{figure}

From the earlier presentation, it is clear that the latent variables $\{\boldsymbol{z}_i|N\geq i \geq 0\}$ form a sequence of progressively noisier versions of the data sample $\boldsymbol{x}$ as index $i$ increases from $0$ to $N$. It is therefore reasonable to assume that as the index $i$ decreases from $N$ until $0$, the estimates $\{\hat{\boldsymbol{x}}( \boldsymbol{z}_i, \hat{\boldsymbol{\epsilon}}_{\boldsymbol{\theta}}(\boldsymbol{z}_i,i))| N\geq i\geq 0\}$ in (
\ref{equ:x_epsilon}) are increasingly accurate. As shown in Fig.~\ref{fig:extraP}, as $i$ decreases, the estimate $\hat{\boldsymbol{x}}( \boldsymbol{z}_i, \hat{\boldsymbol{\epsilon}}_{\boldsymbol{\theta}}(\boldsymbol{z}_i,i))$ becomes increasingly close to $\boldsymbol{x}$. As the Gaussian noise $\boldsymbol{\epsilon}_i$ in $\boldsymbol{z}_i$ is a random variable, the estimate $\hat{\boldsymbol{x}}_{\boldsymbol{\theta}}(\boldsymbol{z}_i, i)$ should also be treated as following a certain distribution. If the model $\hat{\boldsymbol{\epsilon}}_{\boldsymbol{\theta}}$ is well trained, the variances of the estimates should be upper-bounded. By following the above guidelines, we make an assumption to the estimates for $\boldsymbol{x}$ below. We will use the assumption later on to investigate the strengths of the extrapolation introduced in LA-DPMs.

\begin{algorithm}[t!]
\begin{algorithmic}
\caption{Sampling of an LA-DDPM}
\label{alg:LADDPM}
\STATE {\small \textbf{Input:}  $\boldsymbol{z}_N$ and  $\hat{\boldsymbol{x}}( \boldsymbol{z}_{N+1}, \hat{\boldsymbol{\epsilon}}_{\boldsymbol{\theta}}(\boldsymbol{z}_{n+1},N+1))=0$, $\lambda_N=0$} 
\FOR{\small $i=N, \ldots, 1$} 
\STATE {\small Compute $\hat{\boldsymbol{x}}( \boldsymbol{z}_i, \hat{\boldsymbol{\epsilon}}_{\boldsymbol{\theta}}(\boldsymbol{z}_i,i))$}
\STATE {\small $\begin{array}{l}\tilde{\boldsymbol{x}}_{\boldsymbol{\theta},i}(\lambda_i)\hspace{-0.3mm}=\hspace{-0.3mm}(1+\lambda_i)\hat{\boldsymbol{x}}( \boldsymbol{z}_i, \hat{\boldsymbol{\epsilon}}_{\boldsymbol{\theta}}(\boldsymbol{z}_i,i)) \hspace{-0.3mm}\\
\hspace{15mm}-\hspace{-0.3mm} \lambda_i\hat{\boldsymbol{x}}( \boldsymbol{z}_{i+1}, \hat{\boldsymbol{\epsilon}}_{\boldsymbol{\theta}}(\boldsymbol{z}_{i+1},i+1))\end{array}$ }
\STATE {\small $\begin{array}{l}\boldsymbol{\mu}(\boldsymbol{z}_{i-1}|\boldsymbol{z}_i,i,\boldsymbol{z}_{i+1},i+1 ) = \frac{\sigma_{i-1}^2}{\sigma_i^2}\alpha_{i|i-1}\boldsymbol{z}_i\nonumber \\\hspace{20mm}+\frac{\sigma_{i|i-1}^2}{\sigma_i^2}\alpha_{i-1}\tilde{\boldsymbol{x}}_{\boldsymbol{\theta},i}(\lambda_i)\end{array}$  }
\STATE {\small $\boldsymbol{z}_{i-1}=\boldsymbol{\mu}(\boldsymbol{z}_{i-1}|\boldsymbol{z}_i,i,\boldsymbol{z}_{i+1},i+1)+\varphi_i \boldsymbol{\epsilon}$ }
\ENDFOR        \STATE {\small \textbf{output:} $\hat{\boldsymbol{x}}( \boldsymbol{z}_0, \hat{\boldsymbol{\epsilon}}_{\boldsymbol{\theta}}(\boldsymbol{z}_0,0))$ }
\end{algorithmic}
\end{algorithm}

\begin{assumption}
\label{assumption1}
The estimates $\{\hat{\boldsymbol{x}}( \boldsymbol{z}_i, \hat{\boldsymbol{\epsilon}}_{\boldsymbol{\theta}}(\boldsymbol{z}_i,i))| N\geq i\geq 0\}$ are assumed to be represented in terms of $\boldsymbol{x}$ as
\begin{align}
    \hat{\boldsymbol{x}}_{\boldsymbol{\theta}}(\boldsymbol{z}_{j}, j)  &= \gamma_j \boldsymbol{x}+\phi_i\boldsymbol{\epsilon}_{b,j}, 
    \label{equ:esti_1}\end{align}
where $\phi_i<M$, and for simplicity, the residual noise $\boldsymbol{\epsilon}_{b,j}$ is assumed to follow a spherical Gaussian distribution, i.e., $\boldsymbol{\epsilon}_{b,j}\sim \mathcal{N}(0,\boldsymbol{I})$, and for $0\leq j<k\leq N$, we have \begin{align}
1>\gamma_j > \gamma_k\geq 0,\quad 0\leq\varphi_j <\varphi_k. \label{equ:esti_2}  
\end{align}
Furthermore, the estimate $ \hat{\boldsymbol{x}}( \boldsymbol{z}_{i+1}, \hat{\boldsymbol{\epsilon}}_{\boldsymbol{\theta}}(\boldsymbol{z}_{i+1},i+1)) $ can be represented in terms of $ \hat{\boldsymbol{x}}( \boldsymbol{z}_i, \hat{\boldsymbol{\epsilon}}_{\boldsymbol{\theta}}(\boldsymbol{z}_i,i)) $ as
\begin{align}
&\hat{\boldsymbol{x}}( \boldsymbol{z}_{i+1}, \hat{\boldsymbol{\epsilon}}_{\boldsymbol{\theta}}(\boldsymbol{z}_{i+1},i+1)) \nonumber \\ &=\gamma_{i+1|i}\hat{\boldsymbol{x}}( \boldsymbol{z}_i, \hat{\boldsymbol{\epsilon}}_{\boldsymbol{\theta}}(\boldsymbol{z}_i,i))+\phi_{i+1|i}\boldsymbol{\epsilon}_{b,i+1|i}, 
 \label{equ:esti_3}  \end{align} 
where $\gamma_{i+1|i}=\gamma_{i+1}/\gamma_{i}\in (0,1)$, $\phi_{i+1|i}^2=\phi_{i+1}^2-\gamma_{i+1|i}^2\phi_i^2>0$, and $\boldsymbol{\epsilon}_{b,i+1|i}\sim \mathcal{N}(0,\boldsymbol{I})$. That is, the estimates $
    \{\hat{\boldsymbol{x}}( \boldsymbol{z}_j, \hat{\boldsymbol{\epsilon}}_{\boldsymbol{\theta}}(\boldsymbol{z}_j,j)) | N\geq j\geq 0\}$ form a Markov process.   
\label{assum1}
\end{assumption}

        

Next, we briefly consider the two consecutive estimates $\hat{\boldsymbol{x}}( \boldsymbol{z}_i, \hat{\boldsymbol{\epsilon}}_{\boldsymbol{\theta}}(\boldsymbol{z}_i,i))$ and $\hat{\boldsymbol{x}}( \boldsymbol{z}_{i+1}, \hat{\boldsymbol{\epsilon}}_{\boldsymbol{\theta}}(\boldsymbol{z}_{i+1},i+1))$. It is clear from
(\ref{equ:esti_1})-(\ref{equ:esti_2}) that as $j$ decreases from $i+1$ to $i$, the estimate
$\hat{\boldsymbol{x}}( \boldsymbol{z}_j, \hat{\boldsymbol{\epsilon}}_{\boldsymbol{\theta}}(\boldsymbol{z}_j,j))$ becomes more accurate. By applying (\ref{equ:esti_1})-(\ref{equ:esti_3}), the difference of the two estimates can be represented as 
\begin{align}
  \Delta \hat{\boldsymbol{x}}_{\boldsymbol{\theta},i} &=  \hat{\boldsymbol{x}}( \boldsymbol{z}_i, \hat{\boldsymbol{\epsilon}}_{\boldsymbol{\theta}}(\boldsymbol{z}_i,i)) - \hat{\boldsymbol{x}}( \boldsymbol{z}_{i+1}, \hat{\boldsymbol{\epsilon}}_{\boldsymbol{\theta}}(\boldsymbol{z}_{i+1},i+1)) \nonumber \\
    &= (1-\gamma_{i+1|i})\gamma_i\boldsymbol{x}+   (1-\gamma_{i+1|i})\phi_i\boldsymbol{\epsilon}_{b,i} \nonumber \\
    &\hspace{3mm}-\phi_{i+1|i}\boldsymbol{\epsilon}_{b,i+1|i}, 
    \label{equ:dif}
\end{align}
where  $\boldsymbol{\epsilon}_{b,i}$ and $\boldsymbol{\epsilon}_{b,i+1|i}$ are independent variables. Because of the term $(1-\gamma_{i+1|i})\gamma_i\boldsymbol{x}$  in (\ref{equ:dif}), the difference $ \Delta \hat{\boldsymbol{x}}_{\boldsymbol{\theta},i} $ provides additional information about $\boldsymbol{x}$ in comparison to $ \hat{\boldsymbol{x}}( \boldsymbol{z}_i, \hat{\boldsymbol{\epsilon}}_{\boldsymbol{\theta}}(\boldsymbol{z}_i,i))$. As demonstrated in Fig.~\ref{fig:extraP}, $\Delta \hat{\boldsymbol{x}}_{\boldsymbol{\theta},i} $ can be viewed as a noisy vector towards $\boldsymbol{x}$ at timestep $i$. From a high level point of view, it provides additional gradient-descent information that could be exploited to refine the estimate for $\boldsymbol{x}$ at timestep $i$.   

\subsection{LA-DDPM} 
In this subsection, we incorporate the additional gradient information $\Delta \hat{\boldsymbol{x}}_{\boldsymbol{\theta},i}$ of (\ref{equ:dif}) into the backward update expression for $\boldsymbol{z}_{i-1}$ in the DDPM model. In particular, (\ref{equ:Gausian}) is modified to be 
\begin{align}
 &p(\boldsymbol{z}_{i-1}|\boldsymbol{z}_{i:N})   \nonumber \\ 
 &\approx q(\boldsymbol{z}_{i-1}|  \boldsymbol{z}_i,\boldsymbol{x} =\tilde{\boldsymbol{x}}_{\boldsymbol{\theta},i}(\lambda_i)) 
 \notag \\
 &=\hspace{-0.8mm} \mathcal{N}\hspace{-0.8mm}\left(\hspace{-0.8mm}\underbrace{\frac{\sigma_{i-1}^2}{\sigma_i^2}\alpha_{i|i-1}\boldsymbol{z}_i\hspace{-0.8mm}+\hspace{-0.8mm}\frac{\sigma_{i|i-1}^2}{\sigma_i^2}\alpha_{i-1}\tilde{\boldsymbol{x}}_{\boldsymbol{\theta},i}(\lambda_i)}_{\boldsymbol{\mu}(\boldsymbol{z}_{i-1}| \boldsymbol{z}_i, i, \boldsymbol{z}_{i+1},i+1) }, \underbrace{\frac{\sigma_{i-1}^2\sigma_{i|i-1}^2}{\sigma_i^2}}_{\varphi_i}\boldsymbol{I}\hspace{-0.8mm}\right)\hspace{-0.8mm}, \label{equ:Gausian_lookahead1}
\end{align}
where $\tilde{\boldsymbol{x}}_{\boldsymbol{\theta},i}(\lambda_i)$ is computed in the form of
\begin{align}
    &\hspace{0mm}\tilde{\boldsymbol{x}}_{\boldsymbol{\theta},i}(\lambda_i) \nonumber \\
    &= \hat{\boldsymbol{x}}(\boldsymbol{z}_i,\hat{\boldsymbol{\epsilon}}_{\boldsymbol{\theta}}(\boldsymbol{z}_i,i)) + \lambda_i \Delta \hat{\boldsymbol{x}}_{\boldsymbol{\theta},i} \nonumber \\
    &= (1\hspace{-0.3mm}+\hspace{-0.3mm}\lambda_i)\hat{\boldsymbol{x}}(\boldsymbol{z}_i,\hat{\boldsymbol{\epsilon}}_{\boldsymbol{\theta}}(\boldsymbol{z}_i,i)) \hspace{-0.6mm}-\hspace{-0.6mm} \lambda_i \hat{\boldsymbol{x}}(\hat{\boldsymbol{\epsilon}}_{\boldsymbol{\theta}}(\boldsymbol{z}_{i+1},i+1)), \label{equ:extra}
\end{align}
where $\lambda_i\geq 0$ denotes the stepsize for incorporating the difference $\Delta \hat{\boldsymbol{x}}_{\boldsymbol{\theta},i}$, and $\lambda_i=0$ reduces to the original update procedure for DDPM.  It is noted from (\ref{equ:extra}) that the new estimate $\tilde{\boldsymbol{x}}_{\boldsymbol{\theta},i}(\lambda_i)$ is obtained by conducting extrapolation over the two consecutive vectors $\hat{\boldsymbol{x}}(\boldsymbol{z}_i,\hat{\boldsymbol{\epsilon}}_{\boldsymbol{\theta}}(\boldsymbol{z}_i,i))$ and $\hat{\boldsymbol{x}}(\boldsymbol{z}_{i+1},\hat{\boldsymbol{\epsilon}}_{\boldsymbol{\theta}}(\boldsymbol{z}_{i+1},i+1))$. As demonstrated in Fig.~\ref{fig:extraP}, the new estimate $\tilde{\boldsymbol{x}}_{\boldsymbol{\theta},i}(\lambda_i)$ is closer to $\boldsymbol{x}$. Conceptually speaking,  the extrapolation operation allows the backward process to look ahead toward a noisy direction targeting $\boldsymbol{x}$. This improves the estimation accuracy for $\boldsymbol{x}$ when the parameter $\lambda_i$ is properly selected. See Alg.~\ref{alg:LADDPM} for a summary of the sampling procedure of an LA-DDPM.




\subsection{LA-DDIM}
\label{subsec:DDIM}
It is known that DDIM extends DDPM by considering a non-Markov forward process $q(\boldsymbol{z}_N|\boldsymbol{x})\prod_{i=1}^{N}q(\boldsymbol{z}_{i-1}|\boldsymbol{z}_i,\boldsymbol{x})$ while keeping the marginal distribution $q(\boldsymbol{z}_i|\boldsymbol{x})$ the same as that of DDPM. Consequently, in the backward process of DDIM, the latent variable $\boldsymbol{z}_{i-1}$ can be estimated with higher accuracy from $\boldsymbol{z}_i$ than DDPM. Specially, $\boldsymbol{z}_{i-1}$ in DDIM is computed in the form of
\begin{align}
    \boldsymbol{z}_{i-1} \hspace{-0.2mm}=& \hspace{0.7mm} \alpha_{i-1} \underbrace{\left(\frac{\boldsymbol{z}_i \hspace{-0.3mm}-\hspace{-0.3mm} \sigma_i\hat{\boldsymbol{\epsilon}}_{\boldsymbol{\theta}}(\boldsymbol{z}_i, i) }{\alpha_i}\right)}_{\hat{\boldsymbol{x}}( \boldsymbol{z}_i, \hat{\boldsymbol{\epsilon}}_{\boldsymbol{\theta}}(\boldsymbol{z}_i,i)) }\hspace{-0.0mm}+\hspace{0.5mm}\sigma_{i-1}\hat{\boldsymbol{\epsilon}}_{\boldsymbol{\theta}}(\boldsymbol{z}_i, i). \label{equ:DDIM}
\end{align}
It is clear from (\ref{equ:DDIM}) that $\boldsymbol{z}_{i-1}$ can be viewed as a linear combination of $\hat{\boldsymbol{x}}(\boldsymbol{z}_i, \hat{\boldsymbol{\epsilon}}_{\boldsymbol{\theta}}(\boldsymbol{z}_i,i))$ and $\hat{\boldsymbol{\epsilon}}_{\boldsymbol{\theta}}(\boldsymbol{z}_i, i)$.

To obtain the update expression for LA-DDIM, we simply modify (\ref{equ:DDIM}) by replacing $\hat{\boldsymbol{x}}( \boldsymbol{z}_i, \hat{\boldsymbol{\epsilon}}_{\boldsymbol{\theta}}(\boldsymbol{z}_i,i))$ with $\tilde{\boldsymbol{x}}_{\boldsymbol{\theta},i}(\lambda_i)$ in (\ref{equ:extra}), which can be represented as
\begin{align}
    \boldsymbol{z}_{i-1} = \alpha_{i-1} \tilde{\boldsymbol{x}}_{\boldsymbol{\theta},i}(\lambda_i)+\sigma_{i-1}\hat{\boldsymbol{\epsilon}}_{\boldsymbol{\theta}}(\boldsymbol{z}_i, i).
    \label{equ:LA-DDIM}
\end{align}

\subsection{LA-DPM-Solver}
In this subsection, we first briefly explain how DPM-Solver of \cite{Lu22DPM_Solver} is motivated. We then consider incorporating the difference vector $\Delta \hat{\boldsymbol{x}}_{\boldsymbol{\theta},i}$ into the update expressions  of  DPM-Solver.

In \cite{Lu22DPM_Solver}, the authors attempted to solve the following ODE derived from the forward process (\ref{equ:ForwardGau}): 
\begin{align}
    \frac{d \boldsymbol{z}_t }{dt} \hspace{-0.4mm}=\hspace{-0.4mm} f(t)\boldsymbol{z}_t \hspace{-0.4mm}+\hspace{-0.4mm} \frac{g^2(t)}{2\sigma_t}\hat{\boldsymbol{\epsilon}}_{\boldsymbol{\theta}}(\boldsymbol{z}_t,t)\;\;\; \boldsymbol{z}_{T=1}\sim \mathcal{N}(0,\tilde{\sigma}\boldsymbol{I}), 
    \label{equ:ODE}
\end{align}
where $f(t)=\frac{d\log \alpha_t}{dt}$ and $g^2(t)=\frac{d\sigma_t^2}{dt}-2\frac{d\log \alpha_t}{dt}\sigma_t^2$. By applying the  ``variation of constants" formula \cite{Atkinson11ODE} to (\ref{equ:ODE}), the exact solution $\boldsymbol{z}_{i-1}$ given $\boldsymbol{z}_i$ can be represented as 
\begin{align}
\boldsymbol{z}_{{i-1}} &= e^{\int_{t_{i}}^{t_{i-1}} f(\tau)d\tau}\boldsymbol{z}_{i} \nonumber \\
&+\int_{t_{i}}^{t_{i-1}}\left(e^{\int_{t_{i}}^{t_{i-1}}f(r)dr} \frac{g^2(\tau)}{2\sigma_{\tau}} \hat{\boldsymbol{\epsilon}}_{\boldsymbol{\theta}}(\boldsymbol{z}_{\tau},\tau) \right)d\tau, \label{equ:ODESol}
\end{align}
which involves an integration over the predicted Gaussion noise vector $\hat{\boldsymbol{\epsilon}}_{\boldsymbol{\theta}}(\boldsymbol{z}_{\tau},\tau)$. 

The authors of \cite{Lu22DPM_Solver} then propose discrete high-order solutions to approximate the integration in (\ref{equ:ODESol}). Before presenting the solutions, we first introduce two functions.  Let $\lambda_t = \log(\alpha_t/\sigma_t)$ denote the logarithm of the SNR-ratio $\alpha_t/\sigma_t$. $\lambda_t$ is a monotonic decreasing function over time $t$. Therefore, one can also define an inverse function from $\lambda$ to $t$, denoted as $t_{\lambda}(\cdot):\mathbb{R}\rightarrow \mathbb{R}$. Upon introducing the above functions, the update expression for the 2nd order discrete solution (referred to as DPM-Solver-2 in \cite{Lu22DPM_Solver}) takes the form of
\begin{align}
    t_{i-\frac{1}{2}} =& t_{\lambda}(\frac{\lambda_{t_{i-1}}+\lambda_{t_i}}{2}),  \label{equ:solver2_1} \\
    \boldsymbol{z}_{i-\frac{1}{2}} =&  \frac{\alpha_{{i-\frac{1}{2}}}}{\alpha_{i}}\boldsymbol{z}_{i} - \sigma_{i-\frac{1}{2}}(e^{\frac{h_i}{2}}-1)\hat{\boldsymbol{\epsilon}}_{\boldsymbol{\theta}}(\boldsymbol{z}_{i}, i), \label{equ:solver2_2} \\
    \boldsymbol{z}_{i-1} \hspace{-0.8mm}=\hspace{-0.2mm}& \frac{\alpha_{i-1}}{\alpha_{i}}\boldsymbol{z}_{i} \hspace{-0.8mm} -\hspace{-0.8mm} \sigma_{i-1}(e^{h_i}\hspace{-0.8mm}-\hspace{-0.8mm}1)\hat{\boldsymbol{\epsilon}}_{\boldsymbol{\theta}}\left(\boldsymbol{z}_{i-\frac{1}{2}},i-\frac{1}{2}\right) \hspace{-0.5mm}, \label{equ:solver2_3}
\end{align}
where $h_i=\lambda_{t_{i-1}} - \lambda_{t_i}$. The subscript $i-\frac{1}{2}$ indicates that the time $t_{i-\frac{1}{2}}$ is in between $t_{i-1}$ and $t_i$. The latent variable $\boldsymbol{z}_{i-\frac{1}{2}}$ at time $t_{i-\frac{1}{2}}$ is firstly estimated  in preparation for computing $\boldsymbol{z}_{i-1}$. By using the property that $\lambda_{t_{i-\frac{1}{2}}}=(\lambda_{t_{i-1}}+\lambda_{t_{i}})/2$, the update expression (\ref{equ:solver2_2}) for $\boldsymbol{z}_{i-\frac{1}{2}}$ can be simplified to be 
\begin{align}
    \boldsymbol{z}_{i-\frac{1}{2}} \hspace{-0.2mm}=& \hspace{0.6mm} \alpha_{i-\frac{1}{2}} \hat{\boldsymbol{x}}( \boldsymbol{z}_i, \hat{\boldsymbol{\epsilon}}_{\boldsymbol{\theta}}(\boldsymbol{z}_i,i)) \hspace{-0.2mm}+\hspace{-0.2mm}\sigma_{i-\frac{1}{2}}\hat{\boldsymbol{\epsilon}}_{\boldsymbol{\theta}}(\boldsymbol{z}_i, i), \label{equ:Solver2_inter}
\end{align}
which in fact coincides with the update expression (\ref{equ:DDIM}) for DDIM.


We are now in a position to design LA-DPM-Solver-2. Similarly to LA-DDIM, we modify (\ref{equ:Solver2_inter}) by replacing $\hat{\boldsymbol{x}}( \boldsymbol{z}_i, \hat{\boldsymbol{\epsilon}}_{\boldsymbol{\theta}}(\boldsymbol{z}_i,i))$ with an extrapolated term: 
\begin{align}
  \hspace{-3mm}  \boldsymbol{z}_{i-\frac{1}{2}} \hspace{-0.2mm}=& \hspace{0.4mm} \alpha_{i-\frac{1}{2}} \Big[(1+\lambda_i)\hat{\boldsymbol{x}}( \boldsymbol{z}_i, \hat{\boldsymbol{\epsilon}}_{\boldsymbol{\theta}}(\boldsymbol{z}_i,i)) \nonumber \\
    &\hspace{-10mm} - \hspace{-0.6mm} \lambda_i \hat{\boldsymbol{x}}( \boldsymbol{z}_{i+\frac{1}{2}}, \hat{\boldsymbol{\epsilon}}_{\boldsymbol{\theta}}(\boldsymbol{z}_{i+\frac{1}{2}},i\hspace{-0.8mm}+\hspace{-0.8mm}\frac{1}{2}))  \Big] \hspace{-0.2mm}+\hspace{-0.2mm}\sigma_{i-\frac{1}{2}}\hat{\boldsymbol{\epsilon}}_{\boldsymbol{\theta}}(\boldsymbol{z}_i, i). \label{equ:Solver2_inter_LA}
\end{align}
Once $\boldsymbol{z}_{i-\frac{1}{2}}$ is computed,  The computation for $\boldsymbol{z}_{i-1}$ follows directly from (\ref{equ:solver2_3}). 

\begin{remark}
In \cite{Lu22DPM_Solver}, the authors further propose DPM-Solver-3, the 3rd order discrete solution for approximating (\ref{equ:ODESol}). Correspondingly, we propose LA-DPM-Solver-3. See Appendix~\ref{appendix:LA_Solver_3} for the update expressions.    
\end{remark}

\subsection{Analysis of estimation accuracy for $\boldsymbol{x}$}
In this subsection, we derive the optimal setup $\lambda_i^{\ast}$ for $\lambda_i$ under Assumption~\ref{assum1}. Our objective is to find out under what condition, $\lambda_i^{\ast}$ is positive, indicating that the extrapolation operation improves the estimation accuracy for $\boldsymbol{x}$. To do so, we minimize the expected squared error $\|\tilde{\boldsymbol{x}}_{\boldsymbol{\theta},i}(\lambda_i)-\boldsymbol{x}\|)^2$ conditioned on $\boldsymbol{x}$ in terms of $\lambda_i$:  
\begin{align}
    \lambda_i^{\ast}=\arg\min_{\lambda_i} \mathbb{E}[\|\tilde{\boldsymbol{x}}_{\boldsymbol{\theta},i}(\lambda_i)-\boldsymbol{x}\|^2 |\boldsymbol{x}]. \label{equ:lambdaMin1}
\end{align}
By using (\ref{equ:dif})-(\ref{equ:extra}) and the property that $\{\hat{\boldsymbol{x}}_{\boldsymbol{\theta}}(\boldsymbol{z}_j,\hat{\boldsymbol{\epsilon}}_{\boldsymbol{\theta}}(\boldsymbol{z}_j, j))\}$ follows a Gaussian distribution as stated in Assumption~1, (\ref{equ:lambdaMin1}) can be simplified to be 
\begin{align}
    \lambda_i^{\ast}=&\arg\min_{\lambda_i} ((1+\lambda_i-\lambda_i\gamma_{i+1|i})\gamma_i-1)^2\|\boldsymbol{x}\|^2 \nonumber \\
    &+ (1+\lambda_i-\lambda_i\gamma_{i+1|i})^2\phi_i^2+\lambda_i^2\phi_{i+1|i}^2. \label{equ:lambdaMin2}
\end{align}
It is clear that the RHS of (\ref{equ:lambdaMin2}) is a quadratic function of $\lambda_i$.  The optimal solution $\lambda_i^{\ast}$ can be derived easily and can be expressed as 
\begin{align}
\lambda_i^{\ast} =  \frac{(1-\gamma_{i+1|i})(\gamma_i(1-\gamma_i)\|\boldsymbol{x}\|^2-\phi_i^2)}{(1-\gamma_{i+1|i})^2\gamma_i^2\|\boldsymbol{x}\|^2+(1-\gamma_{i+1|i})^2\phi_i^2+\phi_{i+1|i}^2}.
\label{equ:lambdaOpt}
\end{align}

With (\ref{equ:lambdaOpt}), one can obtain the condition that leads to $\lambda_i^{\ast}>0$. We present the results in a proposition below: 
\begin{proposition}
Suppose the conditions for $\{\hat{\boldsymbol{x}}_{\boldsymbol{\theta}}(\boldsymbol{z}_i,\hat{\boldsymbol{\epsilon}}_{\boldsymbol{\theta}}(\boldsymbol{z}_i, i))\}$ in Assumption~\ref{assumption1} hold. The optimal setup $\lambda_i^{\ast}$ is positive (i.e., $\lambda_i^{\ast}>0$) when the noise amplitude $\phi_i$ satisfies the following inequality 
\begin{align}
    \phi_i^2< \gamma_i(1-\gamma_i)\|\boldsymbol{x}\|^2. \label{equ:cond_lambda}
\end{align}
\end{proposition}

The condition (\ref{equ:cond_lambda}) indicates that if the outputs of the DNN model $ \hat{\boldsymbol{\epsilon}}_{\boldsymbol{\theta}}$ are not too noisy, namely $\{\phi_i\}$ are small in comparison to $\|\boldsymbol{x}\|^2$, then it is desirable to apply the extrapolation operation for the purpose of refining the estimate of $\boldsymbol{x}$. In other words, if the model $ \hat{\boldsymbol{\epsilon}}_{\boldsymbol{\theta}}$ is well designed and trained, one should introduce the additional connections in the sampling procedure of a DPM model.  It is noted that the analysis above is based on approximations of Markov Gaussian distributions made in Assumption~\ref{assumption1}. In practice, it is suggested to find the optimal values of $\{\lambda_j^{\ast}\}_{j=0}^N$ by training an additional DNN instead of relying on the expression (\ref{equ:lambdaOpt}).  As will be demonstrated later on, it is found empirically that a constant $\lambda$ value in LA-DPMs leads to significant performance gain over traditional DPMs even though it may not be optimal.   

\section{Advanced Lookahead Diffusion Models}

In this section, we explain how to introduce additional extrapolation into DEIS and S-PNDM. These methods already employ high-order polynomials of the historical estimated Gaussian noises $\{\hat{\boldsymbol{\epsilon}}_{\boldsymbol{\theta}}(\boldsymbol{z}_{i+j},i+j) | r\geq j\geq 0\}$ in the estimation of the latent variable $\boldsymbol{z}_{i-1}$ at iteration $i$.

For simplicity, we first consider extending DEIS to obtain LA-DEIS. By following \cite{Zhang22DEIS}, the update expression for $\boldsymbol{z}_{i-1}$ in the backward process takes the form 
\begin{align}
\boldsymbol{z}_{i-1} = \frac{\alpha_{i-1}}{\alpha_i}\boldsymbol{z}_i + \sum_{j=0}^rc_{ij}\hat{\boldsymbol{\epsilon}}_{\boldsymbol{\theta}}(\boldsymbol{z}_{i+j}, i+j), 
\label{equ:DEIS}
\end{align}
where the $\{c_{ij}\}_{j=0}^r$ are pre-computed hyper-parameters for the purpose of more accurately approximating an integration of the ODE (\ref{equ:ODE}) for (\ref{equ:ForwardGau})-(\ref{equ:ForwardGauAdd}). 

Next, we reformulate (\ref{equ:DEIS}) into an expression similar to (\ref{equ:DDIM}) for DDIM:
\begin{align}
\boldsymbol{z}_{i-1} = \alpha_{i-1}\underbrace{\left(\frac{ \boldsymbol{z}_i - \sigma_i\tilde{\boldsymbol{\epsilon}}_{[i:i+r]}}{\alpha_i}   \right)}_{\ddot{\boldsymbol{x}}_{[i:i+r]}} + \sigma_{i-1} \tilde{\boldsymbol{\epsilon}}_{[i:i+r]}, 
\label{equ:DEIS2}
\end{align}
where $\tilde{\boldsymbol{\epsilon}}_{[i:i+r]}$ is given by
\begin{align}
\hspace{-2mm}\tilde{\boldsymbol{\epsilon}}_{[i:i+r]} =  \sum_{j=0}^rc_{ij}\hat{\boldsymbol{\epsilon}}_{\boldsymbol{\theta}}(\boldsymbol{z}_{i+j}, i+j)/\hspace{-0.3mm}\left(\sigma_{i-1} \hspace{-0.5mm}-\hspace{-0.5mm} \frac{\alpha_{i-1}\sigma_i}{\alpha_i} \right).
\label{equ:DEIS_noise}
\end{align}
The quantity $\tilde{\boldsymbol{\epsilon}}_{[i:i+r]}$ can be viewed as a more accurate estimate of the Gaussian noise than $\hat{\boldsymbol{\epsilon}}_{\boldsymbol{\theta}}(\boldsymbol{z}_i,i)$ in DDIM. With $\tilde{\boldsymbol{\epsilon}}_{[i:i+r]}$, we can compute an estimate $\ddot{\boldsymbol{x}}_{[i:i+r]}$ of the original data sample $\boldsymbol{x}$. 

Upon obtaining (\ref{equ:DEIS2})-(\ref{equ:DEIS_noise}),  we can easily design LA-DEIS by introducing an additional extrapolation into (\ref{equ:DEIS2}), which can be represented by 
\begin{align}
\boldsymbol{z}_{i-1} =& \hspace{0.5mm} \alpha_{i-1}[(1+\lambda_i)\ddot{\boldsymbol{x}}_{[i:i+r]}-\lambda_i \ddot{\boldsymbol{x}}_{[i+1:i+r+1]}]\nonumber\\
&+ \sigma_{i-1}\tilde{\boldsymbol{\epsilon}}_{[i:i+r]},
\end{align}
where the estimate $\ddot{\boldsymbol{x}}_{[i+1:i+r+1]}$ is from the previous timestep $i+1$. Our intention with the additional extrapolation is to provide a better estimate for $\boldsymbol{x}$ at timestep $i$. In principle, the estimate  $\ddot{\boldsymbol{x}}_{[i:i+r]}$ should be less noisy than $\ddot{\boldsymbol{x}}_{[i+1:i+r+1]}$. As a result, the difference of the two vectors would approximately point towards $\boldsymbol{x}$, thus providing additional gradient information in computing $\boldsymbol{z}_{i-1}$. 

Similarly to the design of LA-DEIS, S-PNDM can also be easily extended to obtain LA-S-PNDM. See Appendix~\ref{appendix:LA_S_PNDM} for the details.

\section{Numerical Experiments}
In the 1st experiment, we used as basis variants of DDPM and DDIM in \cite{Bao22DPM_cov} that obtain the optimal covariances of the conditional Gaussian distributions in the backward process by employing additional pre-trained DNN models. We will show that the sampling quality can be significantly improved by introducing the proposed extrapolations in the above methods. We also conduct an ablation study for a particular LA-DDIM that investigates how the parameters $\{\lambda_i\}$ affect the FID scores.  

In the 2nd experiment, we evaluate LA-DEIS and LA-S-PNDM and the corresponding original methods. The evaluation for LA-DPM-Solvers can be found in Appendix~\ref{appendix:LA_DPM_evaluation}.

\subsection{Evaluation of covariance-optimised DPMs}
\label{subsec:exp_DDPM_DDIM}
\noindent\textbf{Experimental setup}:
In this experiment, we evaluated the improved DPM models developed in \cite{Bao22DPM_cov}, which make use of trained neural networks to estimate the optimal covariances of the conditional Gaussian distributions in the backward process. Four types of improved DPM models from \cite{Bao22DPM_cov} were tested, which are NPR-DDPM, SN-DDPM, NPR-DDIM, and SN-DDIM, respectively. The two notations ``NPR" and ``SN" refer to two different approaches for designing DNN models to estimate the optimal covariances under different conditions. 

Similarly to \cite{Bao22DPM_cov}, we conducted experiments over three datasets: CIFAR10, ImageNet64, and CelebA64. For each dataset, two pre-trained models were downloaded from the open-source link\footnote{\url{https://github.com/baofff/Extended-Analytic-DPM}} provided in \cite{Bao22DPM_cov}, one for the ``NPR" approach and the other for the ``SN" approach.

\begin{table*}[t!]
\caption{\small Comparison of FID score for CIFAR10, CelebA64, and ImageNet64. The notation ``LA" stands for ``lookahead", where the associated DPM models are obtained by introducing extrapolation accordingly. \textbf{Lower} is better.  \vspace{0mm} } 
\vspace{-0.5mm}
\label{tab:cifar10}
\centering
\begin{tabular}{|c|| c|c|c|c|c|c|| c|c|c|c|c|c|| c|c|c|c|c|c| }
\hline
\footnotesize{Data sets}
&\multicolumn{6}{|c|}{\footnotesize{CIFAR10}} 
&\multicolumn{6}{|c|}{\footnotesize{CelebA64}} 
&\multicolumn{6}{|c|}{\footnotesize{ImageNet64}} 
\\
\hline
   {\footnotesize Timesteps }
& \hspace{-2mm} \footnotesize{10} \hspace{-2mm}
& \hspace{-2mm}  \footnotesize{25}    \hspace{-2mm} & \hspace{-2mm}  \footnotesize{50} 
\hspace{-2mm} & \hspace{-2mm}   \footnotesize{100}  
\hspace{-2mm} & \hspace{-2mm}  \footnotesize{200}  
\hspace{-2mm} & \hspace{-2mm}  \footnotesize{1000} 
\hspace{-2mm} 
& \hspace{-2mm} \footnotesize{10} \hspace{-2mm}
& \hspace{-2mm}  \footnotesize{25}    \hspace{-2mm} & \hspace{-2mm}  \footnotesize{50} 
\hspace{-2mm} & \hspace{-2mm}   \footnotesize{100}  
\hspace{-2mm} & \hspace{-2mm}  \footnotesize{200}  
\hspace{-2mm} & \hspace{-2mm}  \footnotesize{1000} 
\hspace{-2mm} 
& \hspace{-2mm} \footnotesize{10} \hspace{-2mm}
& \hspace{-2mm}  \footnotesize{25}    \hspace{-2mm} & \hspace{-2mm}  \footnotesize{50} 
\hspace{-2mm} & \hspace{-2mm}   \footnotesize{100}  
\hspace{-2mm} & \hspace{-2mm}  \footnotesize{200}  
\hspace{-2mm} & \hspace{-2mm}  \footnotesize{1000} 
\hspace{-2mm} 
 \\
 \hline
\footnotesize{NPR-DDPM} \hspace{-2mm} &  
\hspace{-2mm}  \footnotesize{32.64} \hspace{-2mm} & \hspace{-2mm} 
\footnotesize{10.48} \hspace{-2mm} & \hspace{-2mm} 
\footnotesize{6.18}    \hspace{-2mm} & \hspace{-2mm}  \footnotesize{4.46} \hspace{-2mm} & \hspace{-2mm}  \footnotesize{3.70} \hspace{-2mm} 
& \hspace{-2mm} \footnotesize{4.04} 
\hspace{-2mm} 
& 
\hspace{-2mm}
\footnotesize{28.32} \hspace{-2mm} & \hspace{-2mm} 
\footnotesize{15.51} \hspace{-2mm} & \hspace{-2mm} 
\footnotesize{10.70}  
\hspace{-2mm} & \hspace{-2mm}  \footnotesize{8.28} 
\hspace{-2mm} & \hspace{-2mm}  \footnotesize{7.01} 
\hspace{-2mm} & \hspace{-2mm} \footnotesize{5.26} 
\hspace{-2mm} 
&\hspace{-2mm} 
\footnotesize{53.22} \hspace{-2mm} & \hspace{-2mm} 
\footnotesize{28.41} \hspace{-2mm} & \hspace{-2mm} 
\footnotesize{21.05}  
\hspace{-2mm} & \hspace{-2mm}  \footnotesize{18.26} 
\hspace{-2mm} & \hspace{-2mm}  \footnotesize{\textbf{16.75}}
& \hspace{-2mm}  \footnotesize{{16.30}}
\hspace{-2mm} 
\\ \hline 
\rowcolor{LightCyan}
\footnotesize{LA-NPR-DDPM}\hspace{-2mm}&\hspace{-2mm}
\footnotesize{\textbf{25.59}} \hspace{-2mm} & \hspace{-2mm} 
\footnotesize{\textbf{8.48}} \hspace{-2mm} & \hspace{-2mm} 
\footnotesize{\textbf{5.28}}  
\hspace{-2mm} & \hspace{-2mm}  \footnotesize{\textbf{4.07}} 
\hspace{-2mm} & \hspace{-2mm}  \footnotesize{\textbf{3.47}} 
\hspace{-2mm} & \hspace{-2mm} 
\footnotesize{\textbf{3.90}}
\hspace{-2mm} & \hspace{-2mm} 
\footnotesize{\textbf{25.08}} \hspace{-2mm} & \hspace{-2mm} 
\footnotesize{\textbf{13.92}} \hspace{-2mm} & \hspace{-2mm} 
\footnotesize{\textbf{9.58}}  
\hspace{-2mm} & \hspace{-2mm}  \footnotesize{\textbf{7.43}} 
\hspace{-2mm} & \hspace{-2mm}  \footnotesize{\textbf{6.32}} 
\hspace{-2mm} & \hspace{-2mm}  \footnotesize{\textbf{5.01}}
\hspace{-2mm} 
& \hspace{-2mm} 
\footnotesize{\textbf{48.71}} \hspace{-2mm} & \hspace{-2mm} 
\footnotesize{\textbf{25.42}} \hspace{-2mm} & \hspace{-2mm} 
\footnotesize{\textbf{20.27}}  
\hspace{-2mm} & \hspace{-2mm}  \footnotesize{\textbf{18.16}} 
\hspace{-2mm} & \hspace{-2mm}  \footnotesize{16.83}
& \hspace{-2mm}  \footnotesize{\textbf{16.27}}
\hspace{-2mm} 
\\ \hline 
\footnotesize{gain (\%)}\hspace{-2mm}&\hspace{-2mm}
\footnotesize{{21.6}} \hspace{-2mm} & \hspace{-2mm} 
\footnotesize{{19.1}} \hspace{-2mm} & \hspace{-2mm} 
\footnotesize{{14.6}}  
\hspace{-2mm} & \hspace{-2mm}  \footnotesize{{8.7}} 
\hspace{-2mm} & \hspace{-2mm}  \footnotesize{{6.2}} 
\hspace{-2mm} & \hspace{-2mm} \footnotesize{{3.5}}
\hspace{-2mm}
& \hspace{-2mm} 
\footnotesize{{11.4}} \hspace{-2mm} & \hspace{-2mm} 
\footnotesize{{10.3}} \hspace{-2mm} & \hspace{-2mm} 
\footnotesize{{10.4}}  
\hspace{-2mm} & \hspace{-2mm}  \footnotesize{{10.3}} 
\hspace{-2mm} & \hspace{-2mm}  \footnotesize{{9.8}} 
\hspace{-2mm} & \hspace{-2mm} \footnotesize{4.75}
\hspace{-2mm}
&\hspace{-2mm} 
\footnotesize{8.5} \hspace{-2mm} & \hspace{-2mm} 
\footnotesize{10.5} \hspace{-2mm} & \hspace{-2mm} 
\footnotesize{3.7}  
\hspace{-2mm} & \hspace{-2mm}  \footnotesize{0.5} 
\hspace{-2mm} & \hspace{-2mm}  \footnotesize{-0.5}
\hspace{-2mm} & \hspace{-2mm}  \footnotesize{0.2}
\hspace{-2mm} 
\\
\hline\hline
\footnotesize{SN-DDPM} \hspace{-2mm} & \hspace{-2mm}  
\footnotesize{23.75} \hspace{-2mm} & \hspace{-2mm} 
\footnotesize{6.88} \hspace{-2mm} & \hspace{-2mm} 
\footnotesize{4.58}  \hspace{-2mm} & \hspace{-2mm}  \footnotesize{3.67} \hspace{-2mm} & \hspace{-2mm}  \footnotesize{3.31} \hspace{-2mm} & \hspace{-2mm}  
\footnotesize{3.65} 
\hspace{-2mm}  & \hspace{-2mm}  
\footnotesize{20.55} \hspace{-2mm} & \hspace{-2mm} 
\footnotesize{11.85} \hspace{-2mm} & \hspace{-2mm} 
\footnotesize{7.58}   \hspace{-2mm} & \hspace{-2mm}  \footnotesize{5.95} \hspace{-2mm} & \hspace{-2mm}  \footnotesize{4.96} \hspace{-2mm} & \hspace{-2mm}  
\footnotesize{4.44} 
\hspace{-2mm} 
& \hspace{-2mm}  
\footnotesize{51.09} \hspace{-2mm} & \hspace{-2mm} 
\footnotesize{27.77} \hspace{-2mm} & \hspace{-2mm} 
\footnotesize{20.65}   \hspace{-2mm} & \hspace{-2mm}  \footnotesize{18.07} \hspace{-2mm} & \hspace{-2mm}  \footnotesize{\textbf{16.70}} \hspace{-2mm}
& \hspace{-2mm}  \footnotesize{{16.30}} \hspace{-2mm}
\\ 
\rowcolor{LightCyan}
\hline \footnotesize{LA-SN-DDPM}\hspace{-2mm} & \hspace{-2mm} 
\footnotesize{\textbf{19.75}} \hspace{-2mm} & \hspace{-2mm} 
\footnotesize{\textbf{5.92}} \hspace{-2mm} & \hspace{-2mm} 
\footnotesize{\textbf{4.31}}  
\hspace{-2mm} & 
\hspace{-2mm}
\footnotesize{\textbf{3.55}} 
\hspace{-2mm} 
& \hspace{-2mm}  \footnotesize{\textbf{3.24}} 
\hspace{-2mm} & \hspace{-2mm} 
\footnotesize{\textbf{3.55}}
\hspace{-2mm}  & \hspace{-2mm} 
\footnotesize{\textbf{17.43}} \hspace{-2mm} & \hspace{-2mm} 
\footnotesize{\textbf{10.08}} \hspace{-2mm} & \hspace{-2mm} 
\footnotesize{\textbf{6.41}}  
\hspace{-2mm} & \hspace{-2mm}  \footnotesize{\textbf{5.12}} 
\hspace{-2mm} & \hspace{-2mm}  \footnotesize{\textbf{4.41}} 
\hspace{-2mm} & \hspace{-2mm} \footnotesize{\textbf{4.21}}
\hspace{-2mm} & \hspace{-2mm} 
\footnotesize{\textbf{46.13}} \hspace{-2mm} & \hspace{-2mm} 
\footnotesize{\textbf{24.67}} \hspace{-2mm} & \hspace{-2mm} 
\footnotesize{\textbf{19.83}}  
\hspace{-2mm} & \hspace{-2mm}  \footnotesize{\textbf{17.95}} 
\hspace{-2mm} & \hspace{-2mm}  \footnotesize{16.76} 
\hspace{-2mm} 
 & \hspace{-2mm}  \footnotesize{\textbf{16.28}} 
\hspace{-2mm} 
\\ \hline 
\footnotesize{gain (\%)}\hspace{-2mm} & \hspace{-2mm} 
\footnotesize{{16.8}} \hspace{-2mm} & \hspace{-2mm} 
\footnotesize{{14.0}} \hspace{-2mm} & \hspace{-2mm} 
\footnotesize{{5.9}}  
\hspace{-2mm} & \hspace{-2mm}  \footnotesize{{3.3}} 
\hspace{-2mm} & \hspace{-2mm}  \footnotesize{{2.1}} 
\hspace{-2mm} & \hspace{-2mm} \footnotesize{{2.7}}
\hspace{-2mm} & \hspace{-2mm} 
\footnotesize{15.2} \hspace{-2mm} & \hspace{-2mm} 
\footnotesize{14.9} \hspace{-2mm} & \hspace{-2mm} 
\footnotesize{15.4}  
\hspace{-2mm} & \hspace{-2mm}  \footnotesize{13.9} 
\hspace{-2mm} & \hspace{-2mm}  \footnotesize{11.1} 
\hspace{-2mm} & \hspace{-2mm} \footnotesize{5.2}
\hspace{-2mm} & \hspace{-2mm} 
\footnotesize{9.7} \hspace{-2mm} & \hspace{-2mm} 
\footnotesize{11.2} \hspace{-2mm} & \hspace{-2mm} 
\footnotesize{4.0}  
\hspace{-2mm} & \hspace{-2mm}  \footnotesize{0.7} 
\hspace{-2mm} & \hspace{-2mm}  \footnotesize{-0.4} 
\hspace{-2mm} & \hspace{-2mm}  \footnotesize{0.1} 
\hspace{-2mm} 
\\
\hline
\hline 
\footnotesize{NPR-DDIM}\hspace{-2mm} & \hspace{-2mm} 
\footnotesize{13.41} \hspace{-2mm} & \hspace{-2mm} 
\footnotesize{5.43} \hspace{-2mm} & \hspace{-2mm} 
\footnotesize{3.99}  
\hspace{-2mm} & \hspace{-2mm}  \footnotesize{3.53} 
\hspace{-2mm} & \hspace{-2mm}  \footnotesize{3.40} 
\hspace{-2mm} & \hspace{-2mm} \footnotesize{3.67}
\hspace{-2mm} & \hspace{-2mm} 
\footnotesize{14.94} \hspace{-2mm} & \hspace{-2mm} 
\footnotesize{9.18} \hspace{-2mm} & \hspace{-2mm} 
\footnotesize{6.17}  
\hspace{-2mm} & \hspace{-2mm}  \footnotesize{4.40} 
\hspace{-2mm} & \hspace{-2mm}  \footnotesize{3.67} 
\hspace{-2mm} & \hspace{-2mm} \footnotesize{3.12}
\hspace{-2mm} & \hspace{-2mm} 
\footnotesize{97.27} \hspace{-2mm} & \hspace{-2mm} 
\footnotesize{28.75} \hspace{-2mm} & \hspace{-2mm} 
\footnotesize{19.79}  
\hspace{-2mm} & \hspace{-2mm}  \footnotesize{\textbf{17.71}} 
\hspace{-2mm} & \hspace{-2mm}  \footnotesize{\textbf{17.15}}
& \hspace{-2mm}  \footnotesize{\textbf{17.59}}
\hspace{-2mm} 
\\
\hline
\rowcolor{LightCyan}
\footnotesize{LA-NPR-DDIM}\hspace{-2mm} & \hspace{-2mm} 
\footnotesize{\textbf{10.74}} \hspace{-2mm} & \hspace{-2mm} 
\footnotesize{\textbf{4.71}} \hspace{-2mm} & \hspace{-2mm} 
\footnotesize{\textbf{3.64}}  
\hspace{-2mm} & \hspace{-2mm}  \footnotesize{\textbf{3.33}} 
\hspace{-2mm} & \hspace{-2mm}  \footnotesize{\textbf{3.29}} 
\hspace{-2mm} & \hspace{-2mm} \footnotesize{\textbf{3.49}}
\hspace{-2mm} & \hspace{-2mm} 
\footnotesize{\textbf{14.25}} \hspace{-2mm} & \hspace{-2mm} 
\footnotesize{\textbf{8.83}} \hspace{-2mm} & \hspace{-2mm} 
\footnotesize{\textbf{5.67}}  
\hspace{-2mm} & \hspace{-2mm}  \footnotesize{\textbf{3.76}} 
\hspace{-2mm} & \hspace{-2mm}  \footnotesize{\textbf{2.95}} 
\hspace{-2mm} & \hspace{-2mm} \footnotesize{\textbf{2.95}}
\hspace{-2mm} & \hspace{-2mm} 
\footnotesize{\textbf{71.98}} \hspace{-2mm} & \hspace{-2mm} 
\footnotesize{\textbf{25.39}} \hspace{-2mm} & \hspace{-2mm} 
\footnotesize{\textbf{19.47}}  
\hspace{-2mm} & \hspace{-2mm}  \footnotesize{18.11} 
\hspace{-2mm} & \hspace{-2mm}  \footnotesize{17.89} 
 & \hspace{-2mm}  \footnotesize{18.41}
\hspace{-2mm} 
\\ \hline
\footnotesize{gain (\%)}\hspace{-2mm} & \hspace{-2mm} 
\footnotesize{{19.9}} \hspace{-2mm} & \hspace{-2mm} 
\footnotesize{{13.3}} \hspace{-2mm} & \hspace{-2mm} 
\footnotesize{{8.8}}  
\hspace{-2mm} & \hspace{-2mm}  \footnotesize{{5.7}} 
\hspace{-2mm} & \hspace{-2mm}  \footnotesize{{3.2}} 
\hspace{-2mm} & \hspace{-2mm} \footnotesize{{4.9}}
\hspace{-2mm} & \hspace{-2mm} 
\footnotesize{{4.6}} \hspace{-2mm} & \hspace{-2mm} 
\footnotesize{{3.8}} \hspace{-2mm} & \hspace{-2mm} 
\footnotesize{{8.1}}  
\hspace{-2mm} & \hspace{-2mm}  \footnotesize{{14.5}} 
\hspace{-2mm} & \hspace{-2mm}  \footnotesize{{19.61}} 
\hspace{-2mm} & \hspace{-2mm} \footnotesize{5.4}
\hspace{-2mm} & \hspace{-2mm} 
\footnotesize{26.0} \hspace{-2mm} & \hspace{-2mm} 
\footnotesize{11.7} \hspace{-2mm} & \hspace{-2mm} 
\footnotesize{1.6}  
\hspace{-2mm} & \hspace{-2mm}  \footnotesize{-2.3} 
\hspace{-2mm} & \hspace{-2mm}  \footnotesize{-4.3} 
\hspace{-2mm} & \hspace{-2mm}  \footnotesize{-4.7} 
\hspace{-2mm} 

\\
\hline\hline
\footnotesize{SN-DDIM}\hspace{-2mm} & \hspace{-2mm}
\footnotesize{12.19} \hspace{-2mm} & \hspace{-2mm} 
\footnotesize{4.28} \hspace{-2mm} & \hspace{-2mm} 
\footnotesize{3.39}  
\hspace{-2mm} & \hspace{-2mm}  \footnotesize{3.22} 
\hspace{-2mm} & \hspace{-2mm}  \footnotesize{{4.22}}
\hspace{-2mm} & \hspace{-2mm} \footnotesize{{3.65}}
\hspace{-2mm} & \hspace{-2mm} 
\footnotesize{10.17} \hspace{-2mm} & \hspace{-2mm} 
\footnotesize{5.62} \hspace{-2mm} & \hspace{-2mm} 
\footnotesize{3.90}  
\hspace{-2mm} & \hspace{-2mm}  \footnotesize{3.21} 
\hspace{-2mm} & \hspace{-2mm}  \footnotesize{2.94} 
\hspace{-2mm} & \hspace{-2mm} \footnotesize{2.84} 
\hspace{-2mm} & \hspace{-2mm} 
\footnotesize{91.29} \hspace{-2mm} & \hspace{-2mm} 
\footnotesize{27.74} \hspace{-2mm} & \hspace{-2mm} 
\footnotesize{19.51}  
\hspace{-2mm} & \hspace{-2mm}  \footnotesize{\textbf{17.67}} 
\hspace{-2mm} & \hspace{-2mm}  \footnotesize{\textbf{17.14}} 
\hspace{-2mm}
& \hspace{-2mm}  \footnotesize{\textbf{17.60}} 
\hspace{-2mm}
\\
\hline
\rowcolor{LightCyan}
\footnotesize{LA-SN-DDIM}\hspace{-2mm} & \hspace{-2mm}
\footnotesize{\textbf{8.48}} \hspace{-2mm} & \hspace{-2mm} 
\footnotesize{\textbf{3.15}} \hspace{-2mm} & \hspace{-2mm} 
\footnotesize{\textbf{2.93}}  
\hspace{-2mm} & \hspace{-2mm}  \footnotesize{\textbf{2.92}} 
\hspace{-2mm} & \hspace{-2mm}  \footnotesize{\textbf{3.08}} 
\hspace{-2mm} & \hspace{-2mm} \footnotesize{\textbf{3.47}} \hspace{-2mm} & \hspace{-2mm} 
\footnotesize{\textbf{8.05}} \hspace{-2mm} & \hspace{-2mm} 
\footnotesize{\textbf{4.56}} \hspace{-2mm} & \hspace{-2mm} 
\footnotesize{\textbf{2.93}}  
\hspace{-2mm} & \hspace{-2mm}  \footnotesize{\textbf{2.39}} 
\hspace{-2mm} & \hspace{-2mm}  \footnotesize{\textbf{2.19}} 
\hspace{-2mm} & \hspace{-2mm} \footnotesize{\textbf{2.48}}
\hspace{-2mm} & \hspace{-2mm} 
\footnotesize{\textbf{69.11}} \hspace{-2mm} & \hspace{-2mm} 
\footnotesize{\textbf{25.07}} \hspace{-2mm} & \hspace{-2mm} 
\footnotesize{\textbf{19.32}}  
\hspace{-2mm} & \hspace{-2mm}  \footnotesize{18.06} 
\hspace{-2mm} & \hspace{-2mm}  \footnotesize{17.89} 
\hspace{-2mm}
& \hspace{-2mm}  \footnotesize{18.57} 
\hspace{-2mm}
\\
\hline
\footnotesize{gain (\%)}\hspace{-2mm} & \hspace{-2mm}
\footnotesize{{30.4}} \hspace{-2mm} & \hspace{-2mm} 
\footnotesize{{26.4}} \hspace{-2mm} & \hspace{-2mm} 
\footnotesize{{13.6}}  
\hspace{-2mm} & \hspace{-2mm}  \footnotesize{{9.3}} 
\hspace{-2mm} & \hspace{-2mm}  \footnotesize{{27.0}} 
\hspace{-2mm} & \hspace{-2mm} \footnotesize{{4.9}}\hspace{-2mm} & \hspace{-2mm} 
\footnotesize{{20.8}} \hspace{-2mm} & \hspace{-2mm} 
\footnotesize{{18.9}} \hspace{-2mm} & \hspace{-2mm} 
\footnotesize{{24.9}}  
\hspace{-2mm} & \hspace{-2mm}  \footnotesize{{25.5}} 
\hspace{-2mm} & \hspace{-2mm}  \footnotesize{{25.5}} 
\hspace{-2mm} & \hspace{-2mm} \footnotesize{12.7} \hspace{-2mm} & \hspace{-2mm} 
\footnotesize{24.3} \hspace{-2mm} & \hspace{-2mm} 
\footnotesize{9.6} \hspace{-2mm} & \hspace{-2mm} 
\footnotesize{9.7}  
\hspace{-2mm} & \hspace{-2mm}  \footnotesize{-2.2} 
\hspace{-2mm} & \hspace{-2mm}  \footnotesize{-4.4} 
\hspace{-2mm} & \hspace{-2mm}  \footnotesize{-5.5} 
\hspace{-2mm} 
\\
\hline
\end{tabular}
\vspace{-0mm}
\end{table*}


In the experiment, the strengths of the extrapolations were set to $\lambda_i=0.1$ for all $i\in\{0,1,\ldots, N-1\}$ in all pre-trained models. The tested timesteps for the three datasets were set to $\{10, 25, 50, 100, 200, 1000\}$. For each configuration, 50K artificial images were generated for the computation of FID score.

\noindent\textbf{Performance comparison:} 
The sampling qualities for the three datasets are summarized in  Table~\ref{tab:cifar10}. It is clear that for CIFAR10 and Celeba64, the LA-DPM models outperform the original DPM models significantly for both small and large numbers of timesteps. Roughly speaking, as the number of timesteps decreases from 1000 to 10, the performance gain of LA-DPM increases. That is, it is more preferable to introduce the extrapolation operations when sampling with a limited computational budget. This might be because for a large number of timesteps, the original DPM models are able to maximally exploit the gradient information provided by the DNN model $ \hat{\boldsymbol{\epsilon}}_{\boldsymbol{\theta}}$ and generate high quality samples accordingly. On the other hand, with a small number of timesteps, limited access to the DNN model makes it difficult for the original PDM models to acquire detailed structural information of the data sample $\boldsymbol{x}$. As a result, for a small number of timesteps, the proposed extrapolation operation plays a more important role by improving the mean estimation of the backward conditional Gaussian distributions in the sampling procedure.  

Next, we consider the results obtained for ImageNet64. As shown in Table~\ref{tab:cifar10}, the introduction of extrapolation operations leads to better performance only for a small number of steps (e.g., 10, 25, 50). When the number of steps is large, we observe slightly degraded performance. This is because ImageNet64 is a very large dataset that covers many classes of objects compared to CIFAR10 and CelebA64. As a result, the estimate $\hat{\boldsymbol{x}}_{\boldsymbol{\theta}}(\boldsymbol{z}_j, \hat{\boldsymbol{\epsilon}}_{\boldsymbol{\theta}}(\boldsymbol{z}_j, j))$ may be noisier than the corresponding estimates over CIFAR10 and CelebA. In other words, the setups $\{\lambda_i=0.1\}_{i=0}^{N-1}$ are not appropriate for ImageNet64. In this case, one can simply reduce the strengths (i.e., $\lambda_i\downarrow$) of the extrapolation operations.

\subsection{Ablation study of LA-SN-DDIM over CIFAR10}
In subsection~\ref{subsec:exp_DDPM_DDIM}, the strengths of the extrapolations in LA-SN-DDIM were set to $\{\lambda_i=\lambda=0.1\}_{i=0}^{N-1}$, which led to significant performance gain for small numbers of stepsizes in comparison to SN-DDIM. We now consider how the FID scores change for different setups of $\lambda\in \{0, 0.05, 0.1, 0.15, 0.2, 0.3, 0.4\}$ over timesteps of 10 and 20, where $\{\lambda_i=\lambda\}_{i=0}^{N-1}$ for each $\lambda$ value. 

\begin{figure}[t!]
\centering
\includegraphics[width=60mm]{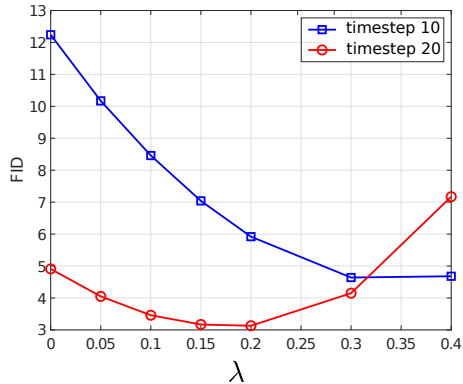}
\vspace*{-0.2cm}
\caption{\small{ FID scores versus $\lambda$ values for LA-SN-DDIM over CIFAR10.}}
\label{fig:FID_lambda}
\vspace{-0.3cm}
\end{figure}

Fig.~\ref{fig:FID_lambda} displays two FID curves over different $\lambda$ values, one for timestep 10 and the other for timestep 20. It is clear that for timestep 10, FID score of around 4.65 can be achieved when $\lambda=0.3$. On the other hand, for timestep 20, FID score of around 3.1 can be achieved when $\lambda=0.2$. This suggests that the setup $\lambda=0.1$ in the first experiment is far from optimality for a small number of timesteps. In other words, the FID scores in Table~\ref{tab:cifar10} can be improved significantly if the $\lambda$ value is tuned for different timesteps.

 \subsection{Evaluation of LA-DEIS and LA-S-PNDM} 
 \label{subsec:DEIS_PNDM}
 \noindent\textbf{Experimental setup}: As noted earlier, DEIS and S-PNDM exploit high-order polynomials of the estimated Gaussian noises per timestep in the backward process for better sampling quality. In this experiment, we demonstrate that their sampling performance can be further improved by introducing additional extrapolation on the estimates of $\boldsymbol{x}$. 
 
 We note that the authors of \cite{Zhang22DEIS} proposed different versions of DEIS depending on how the parameters $\{c_{ij}\}_{j=i}^r$ in (\ref{equ:DEIS}) are computed. For our experiments, we used tAB-DEIS and our new method LA-tAB-DEIS. Furthermore, we also evaluated  S-PNDM and LA-S-PNDM (see the update procedure of Alg.~\ref{alg:LA-S-PNDM} in Appendix~\ref{appendix:LA_S_PNDM}).

The tested pre-trained models are summarized in Table~\ref{tab:sampling_model} in Appendix~\ref{appendix:sampling_model}. In particular, we evaluated LA-tAB-DEIS by utilizing a pre-trained model of VPSDE for CIFAR10 in \cite{Song21SDE_gen}. On the other hand, LA-S-PNDM was evaluated by utilizing four pre-trained models over four popular datasets in \cite{Liu22PNDM}. The tested timesteps for each sampling method are within the range of $[10, 40]$. 

 \begin{figure}[t!]
\centering
\includegraphics[width=85mm]{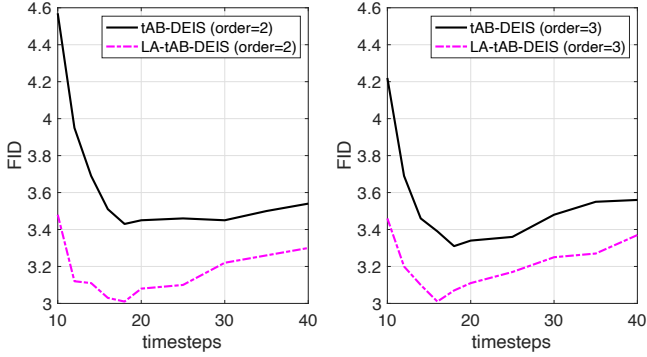}
\vspace*{-0.4cm}
\caption{\small{Performance of tAB-DEIS and LA-tAB-DEIS in terms of FID scores versus timesteps over CIFAR10. The two subplots are for polynomials of order $r=2$ and $r=3$ in (\ref{equ:DEIS}), respectively. The setup $\lambda=0.1$ was employed in LA-tAB-DEIS. } }
\label{fig:DEIS_compare}
\vspace*{-0.3cm}
\end{figure}

\noindent\textbf{Performance comparison}: Fig.~\ref{fig:DEIS_compare} visualizes the FID scores versus tested timesteps for tAB-DEIS and LA-tAB-DEIS. It is clear from this figure that the introduction of additional extrapolation on the estimates of $\boldsymbol{x}$ significantly improves the sampling quality of tAB-DEIS for polynomials of both order $r=2$ and $r=3$.  Similarly to the gain in Table~\ref{tab:cifar10}, the performance gain in Fig.~\ref{fig:DEIS_compare} is relatively large for small timesteps, which is desirable for practical applications. 

\begin{figure}[t!]
\centering
\includegraphics[width=86mm]{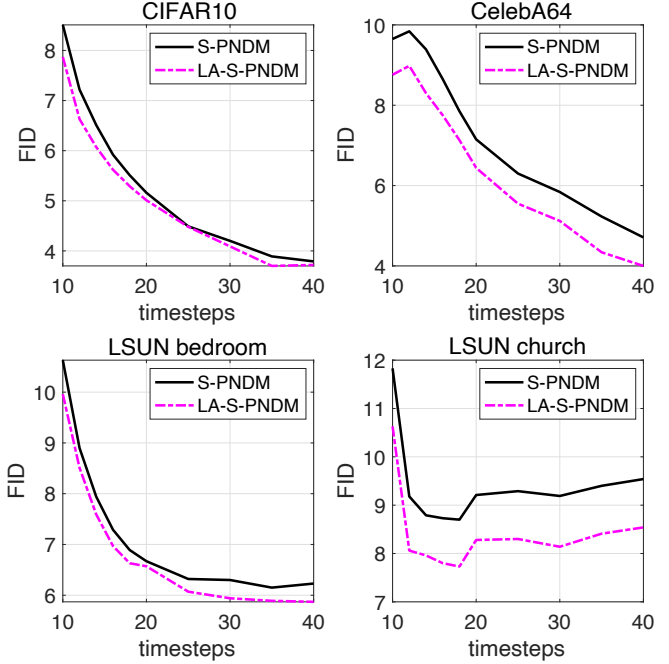}
\vspace*{-0.4cm}
\caption{\footnotesize{ Performance of S-PNDM and LA-S-PNDM over 4 different datasets. The parameter $\lambda$ in LA-S-PNDM was set to $\lambda=0.1$ for $\{\textrm{CIFAR10}, \textrm{CelebA64},\textrm{LSUN church}\}$ and $\lambda=0.05$ for LSUN bedroom. }}
\label{fig:SPNDM}
\vspace{-0.3cm}
\end{figure}

The performance of S-PNDM and LA-S-PNDM is summarized in the four subplots of Fig.~\ref{fig:SPNDM}, one subplot per dataset. It is seen that LA-S-PNDM outperforms S-PNDM consistently over different timesteps and across different datasets, which is consistent with the results of Table~\ref{tab:cifar10} in the 1st experiment. It can also be seen from the figure that the performance gain is more significant for CelebA64 and LSUN church than for CIFAR10 and LSUN bedroom. This might be because different DNN models have different fitting errors when they are being trained.

The above positive results indicate that extrapolation on the estimates of $\boldsymbol{x}$ and the high-order polynomials of the estimated Gaussian noises are compatible. In practice, one should incorporate both techniques in the sampling procedure of DPMs.

\begin{remark}
Due to limited space, we put the experimental results for LA-DPM-Solver-2 and  LA-DPM-Solver-3 in Appendix~\ref{appendix:LA_DPM_evaluation}.
\end{remark}

 \vspace{-3mm}
\section{Conclusions}
 \vspace{-1mm}
In this paper, we proposed a simple approach for improving the estimation accuracy of the mean vectors of a sequence of conditional Gaussian distributions in the backward process of a DPM. A typical DPM model (even including high-order ODE solvers like DEIS and PNDM) first makes a prediction $\hat{\boldsymbol{x}}$ of the data sample $\boldsymbol{x}$ at each timestep $i$, and then uses it in the computation of the mean vector for $\boldsymbol{z}_{i-1}$.  We propose to perform extrapolation on the two most recent estimates of $\boldsymbol{x}$ obtained at times $i$ and $i+1$. In principle, the difference vector of the two estimates approximately points towards $\boldsymbol{x}$, thus providing certain type of gradient information. The extrapolation makes use of the gradient information to obtain a more accurate estimation of $\boldsymbol{x}$, thus improving the estimation accuracy for $\boldsymbol{z}_{i-1}$. 


Extensive experiments showed that the extrapolation operation improves the sampling qualities of variants of DDPM and DDIM, DEIS, S-PNDM, and high-order DPM solvers. It was found that the performance gain is generally more significant for a small number of timesteps. This makes the new technique particularly attractive for settings  with limited computational resources.   




\clearpage
\appendix

\newpage
\onecolumn

\section{Investigation of Computational Overhead}

In addition to the evaluation of sampling qualities, we also examined the computational overhead introduced by the extrapolation operations in LA-DPMs. The minibatch size in the sampling procedure was set to 500, and the running time was measured over a windows machine with a 1080ti GPU.  Table~\ref{tab:cifar10TC} displays the time complexities of different generative models over CIFAR10. It is clear from the table that the computational overhead of the extrapolation operations in LA-DPMs is negligible. This is because the extrapolation operations are linear and no additional DNN models are introduced to assist the operations. 

 \begin{table}[h!]
\caption{\small Comparison of computational costs (measured in units of seconds per minibatch) for CIFAR10.   \vspace{0mm} } 
\label{tab:cifar10TC}
\centering
\begin{tabular}{|c| c|c| c|c| c|c| }
\hline
   {\footnotesize Timesteps }
& \hspace{-2mm} \footnotesize{10} \hspace{-2mm}
& \hspace{-2mm}  \footnotesize{25}    \hspace{-2mm} & \hspace{-2mm}  \footnotesize{50} 
\hspace{-2mm} & \hspace{-2mm}   \footnotesize{100}  
\hspace{-2mm} & \hspace{-2mm}  \footnotesize{200}  
\hspace{-2mm} & \hspace{-2mm}  \footnotesize{1000} 
 \\
 \hline
\footnotesize{NPR-DDPM} \hspace{-2mm} &  
\hspace{-2mm}  \footnotesize{14.9} \hspace{-2mm} & \hspace{-2mm} 
\footnotesize{36.4} \hspace{-2mm} & \hspace{-2mm} 
\footnotesize{70.9} \hspace{-2mm} & \hspace{-2mm}  \footnotesize{139.6} \hspace{-2mm} & \hspace{-2mm}  \footnotesize{278.1} \hspace{-2mm} & \hspace{-2mm}  
\footnotesize{1388.5} 
\hspace{-2mm} 
\\ \hline 
\footnotesize{LA-NPR-DDPM}\hspace{-2mm}&\hspace{-2mm}
\footnotesize{{15.3}} \hspace{-2mm} & \hspace{-2mm} 
\footnotesize{{36.9}} \hspace{-2mm} & \hspace{-2mm} 
\footnotesize{71.6}  
\hspace{-2mm} & \hspace{-2mm}  \footnotesize{{139.8}} 
\hspace{-2mm} & \hspace{-2mm}  \footnotesize{{279.9 }} 
\hspace{-2mm} & \hspace{-2mm} \footnotesize{{1389.9}}
\hspace{-2mm} 
\\
\hline\hline
\footnotesize{SN-DDPM} \hspace{-2mm} & \hspace{-2mm}  
\footnotesize{14.9} \hspace{-2mm} & \hspace{-2mm} 
\footnotesize{36.5} \hspace{-2mm} & \hspace{-2mm} 
\footnotesize{71.0}  \hspace{-2mm} & \hspace{-2mm}  \footnotesize{140.1} \hspace{-2mm} & \hspace{-2mm}  \footnotesize{278.0} \hspace{-2mm} & \hspace{-3mm}  
\footnotesize{1387.9} 
\hspace{-3mm} 
\\ \hline \footnotesize{LA-SN-DDPM}\hspace{-2mm} & \hspace{-2mm} 
\footnotesize{{15.3}} \hspace{-2mm} & \hspace{-2mm} 
\footnotesize{{37.5}} \hspace{-2mm} & \hspace{-2mm}
\footnotesize{{71.9}}  
\hspace{-2mm} & \hspace{-2mm}  \footnotesize{{140.8}} 
\hspace{-2mm} & \hspace{-2mm}  \footnotesize{{278.4}} 
\hspace{-2mm} & \hspace{-2mm} \footnotesize{{1390.7}}
\hspace{-2mm}
\\
\hline
\hline 
\footnotesize{NPR-DDIM}\hspace{-2mm} & \hspace{-2mm} 
\footnotesize{14.9} \hspace{-2mm} & \hspace{-2mm} 
\footnotesize{36.2} \hspace{-2mm} & \hspace{-2mm} 
\footnotesize{70.7}  
\hspace{-2mm} & \hspace{-2mm}  \footnotesize{139.7} 
\hspace{-2mm} & \hspace{-2mm}  \footnotesize{270.2} 
\hspace{-2mm} & \hspace{-2mm} \footnotesize{1385.4}
\hspace{-2mm} 
\\
\hline
\footnotesize{LA-NPR-DDIM}\hspace{-2mm} & \hspace{-2mm} 
\footnotesize{{15.4}} \hspace{-2mm} & \hspace{-2mm} 
\footnotesize{{36.3}} \hspace{-2mm} & \hspace{-2mm} 
\footnotesize{{71.5}}  
\hspace{-2mm} & \hspace{-2mm}  \footnotesize{{140.3}} 
\hspace{-2mm} & \hspace{-2mm}  \footnotesize{{271.1}} 
\hspace{-2mm} & \hspace{-2mm} \footnotesize{{1388.5}}
\hspace{-2mm} 
\\
\hline\hline
\footnotesize{SN-DDIM}\hspace{-2mm} & \hspace{-2mm}
\footnotesize{15.4} \hspace{-2mm} & \hspace{-2mm} 
\footnotesize{36.6} \hspace{-2mm} & \hspace{-2mm} 
\footnotesize{71.1}  
\hspace{-2mm} & \hspace{-2mm}  \footnotesize{136.2} 
\hspace{-2mm} & \hspace{-2mm}  \footnotesize{{279.1}}
\hspace{-2mm} & \hspace{-2mm} \footnotesize{{1386.2}}
\hspace{-2mm} 
\\
\hline
\footnotesize{LA-SN-DDIM}\hspace{-2mm} & \hspace{-2mm}
\footnotesize{{15.6}} \hspace{-2mm} & \hspace{-2mm} 
\footnotesize{{37.0}} \hspace{-2mm} & \hspace{-2mm} 
\footnotesize{{71.3}}  
\hspace{-2mm} & \hspace{-2mm}  \footnotesize{{139.3}} 
\hspace{-2mm} & \hspace{-2mm}  \footnotesize{{280.9}} 
\hspace{-2mm} & \hspace{-2mm} \footnotesize{{1391.3}}
\hspace{-2mm} 
\\
\hline
\end{tabular}
\vspace{-0mm}
\end{table}
 

\section{Additional Ablation Study of LA-DPMs}

We have conducted additional ablation studies for LA-DPMs over both CIFAR10 and ImageNet64. Our main objective is to show that the optimal setup for the parameter $\lambda$ is different for different timesteps.  

\begin{figure}[h!]
\centering
\includegraphics[width=120mm]{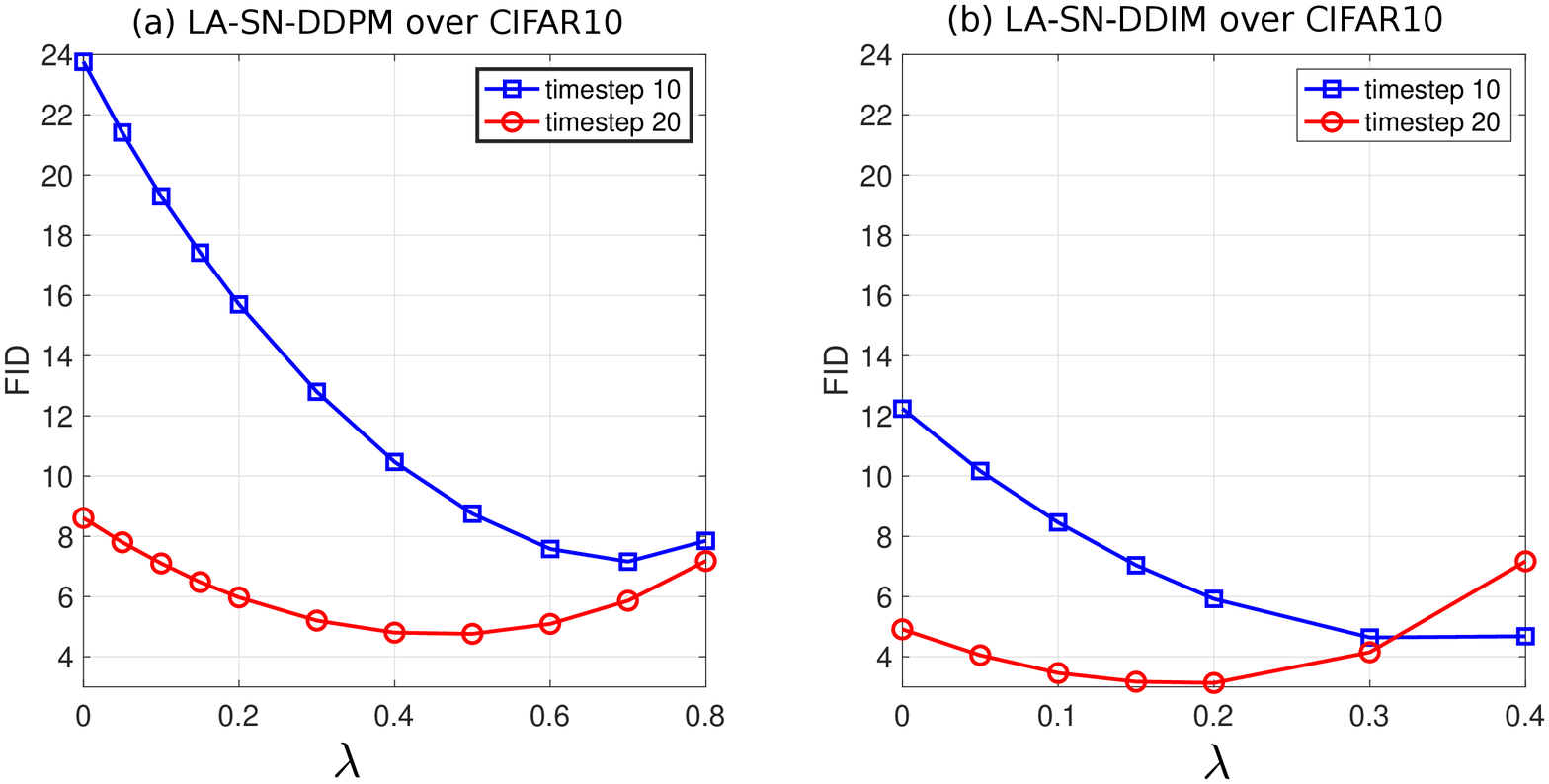}
\vspace*{-0.2cm}
\caption{\footnotesize{ FID scores versus $\lambda$ values for LA-SN-DDPM and LA-SN-DDIM over CIFAR10. When $\lambda=0$, LA-SN-DDPM reduces to SN-DDPM and LA-SN-DDIM reduces to SN-DDIM. }}
\label{fig:FID_SN_CIFAR10}
\vspace{-0.0cm}
\end{figure}

\begin{figure}[h!]
\centering
\includegraphics[width=80mm]{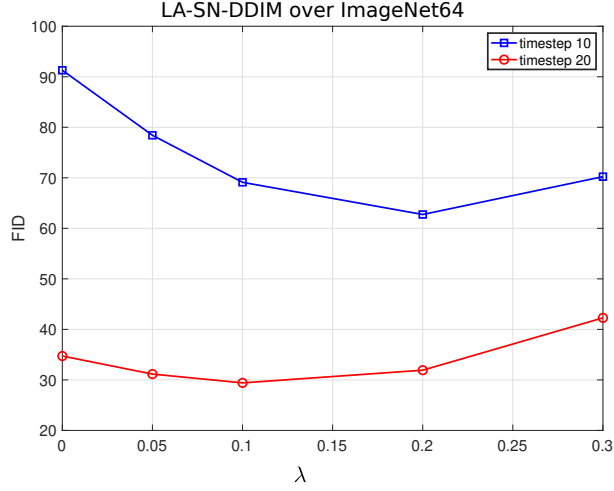}
\vspace*{-0.2cm}
\caption{\footnotesize{ FID scores versus $\lambda$ values for LA-SN-DDIM over ImageNet64. When $\lambda=0$, LA-SN-DDIM reduces to SN-DDIM. }}
\label{fig:FID_SN_ImageNet}
\vspace{-0.0cm}
\end{figure}

It is seen from both Fig.~\ref{fig:FID_SN_CIFAR10} and \ref{fig:FID_SN_ImageNet} that as $\lambda$ increases, the FID score first decreases then increases. That is, it is preferable to select a proper nonzero $\lambda$ to achieve small FID scores.  Furthermore, as the timestep increases from 10 to 20, the optimal value for $\lambda$ decreases. In other words, large $\lambda$ values are preferable when the timestep for sampling is small. This also explains why the setup $\lambda=0.1$ leads to higher FID scores for ImageNet64 for large timesteps (e.g., 200, 1000) in Table~\ref{tab:cifar10}.   

\section{Lookahead High-Order DPM Solvers and Performance Comparison}

\subsection{LA-DPM-Solver-2}

The update expression for DPM-Solver-2 takes the form of (see \cite{Lu22DPM_Solver}) 
\begin{align}
   \left\{\begin{array}{l} t_{i-\frac{1}{2}} = t_{\lambda}(\frac{\lambda_{t_{i-1}}+\lambda_{t_i}}{2})  \\
    \boldsymbol{z}_{i-\frac{1}{2}} =  \frac{\alpha_{{i-\frac{1}{2}}}}{\alpha_{i}}\boldsymbol{z}_{i} - \sigma_{i-\frac{1}{2}}(e^{\frac{h_i}{2}}-1)\hat{\boldsymbol{\epsilon}}_{\boldsymbol{\theta}}(\boldsymbol{z}_{i}, i) \\
    \boldsymbol{z}_{i-1} \hspace{-0.8mm}=\hspace{-0.2mm} \frac{\alpha_{i-1}}{\alpha_{i}}\boldsymbol{z}_{i} \hspace{-0.8mm} -\hspace{-0.8mm} \sigma_{i-1}(e^{h_i}\hspace{-0.8mm}-\hspace{-0.8mm}1)\hat{\boldsymbol{\epsilon}}_{\boldsymbol{\theta}}\left(\boldsymbol{z}_i,i\right) \hspace{-0.6mm}-\hspace{-0.8mm} \sigma_{i-1}(e^{h_i}\hspace{-0.8mm}-\hspace{-0.8mm}1)\hspace{-1mm}\left[ \hat{\boldsymbol{\epsilon}}_{\boldsymbol{\theta}}\left(\boldsymbol{z}_{i-\frac{1}{2}},i\hspace{-0.8mm}-\hspace{-0.8mm}\frac{1}{2}\right)\hspace{-0.8mm}-\hspace{-0.8mm} \hat{\boldsymbol{\epsilon}}_{\boldsymbol{\theta}}\left(\boldsymbol{z}_i,i\right)   \right] \end{array}\right., \label{equ:solver2}
\end{align}
where $\lambda_t =\log (\alpha_t/\sigma_t)$ is a strictly decreasing function and $t_{\lambda}(\cdot)$ is the reverse function of $\lambda_t$, and $h_i = \lambda_{t_{i-1}} - \lambda_{t_i}$. The expression for $\boldsymbol{z}_{i-\frac{1}{2}}$ in (\ref{equ:solver2}) can be simplified to be 
\begin{align}
\boldsymbol{z}_{i-\frac{1}{2}} &=  \frac{\alpha_{{i-\frac{1}{2}}}}{\alpha_{i}}\boldsymbol{z}_{i} - \sigma_{i-\frac{1}{2}}(e^{\frac{h_i}{2}}-1)\hat{\boldsymbol{\epsilon}}_{\boldsymbol{\theta}}(\boldsymbol{z}_{i}, i) \nonumber \\
&\textcolor{blue}{\left[\lambda_{t_{i-\frac{1}{2}}} = \frac{\lambda_{t_{i-1}}+\lambda_{t_i}}{2},\textrm{which is obtained from definition of } t_{i-\frac{1}{2}} \textrm{ in } (\ref{equ:solver2}) \right]} \nonumber \\
&= \frac{\alpha_{{i-\frac{1}{2}}}}{\alpha_{i}}\boldsymbol{z}_{i} - \sigma_{i-\frac{1}{2}}\left(e^{\left(\lambda_{t_{i-\frac{1}{2}}}-\lambda_{t_i} \right) }-1\right)\hat{\boldsymbol{\epsilon}}_{\boldsymbol{\theta}}(\boldsymbol{z}_{i}, i) \nonumber \\  
&\textcolor{blue}{\left[\lambda_{t_{i-\frac{1}{2}}} = \frac{\lambda_{t_{i-1}}+\lambda_{t_i}}{2}=\log(\alpha_{i-\frac{1}{2}}/\sigma_{i-\frac{1}{2}}),\hspace{2mm} \lambda_{t_i} =\log(\alpha_i/\sigma_i) \right]} \nonumber \\
&= \frac{\alpha_{{i-\frac{1}{2}}}}{\alpha_{i}}\boldsymbol{z}_{i} - \sigma_{i-\frac{1}{2}}\left(\frac{\alpha_{i-\frac{1}{2}}}{\sigma_{i-\frac{1}{2}}}\frac{\sigma_{i}}{\alpha_{i}}  -1\right)\hat{\boldsymbol{\epsilon}}_{\boldsymbol{\theta}}(\boldsymbol{z}_{i}, i) \nonumber \\  
&= \alpha_{{i-\frac{1}{2}}}\underbrace{\left(\frac{\boldsymbol{z}_{i}}{ \alpha_{i}} - \frac{\sigma_i}{\alpha_i} \hat{\boldsymbol{\epsilon}}_{\boldsymbol{\theta}}(\boldsymbol{z}_{i}, i)\right)}_{\hat{\boldsymbol{x}}(\boldsymbol{z}_i,\hat{\boldsymbol{\epsilon}}_{\boldsymbol{\theta}}(\boldsymbol{z}_i,i))} + \sigma_{i-\frac{1}{2}}\hat{\boldsymbol{\epsilon}}_{\boldsymbol{\theta}}(\boldsymbol{z}_{i}, i) \nonumber \\
&\textcolor{blue}{\left[\textrm{For variance preserving process: }\sigma_{i-\frac{1}{2}} =\sqrt{1-\alpha_{i-\frac{1}{2}}^2}\right] } \nonumber \\
&= \alpha_{{i-\frac{1}{2}}}\hat{\boldsymbol{x}}(\boldsymbol{z}_i,\hat{\boldsymbol{\epsilon}}_{\boldsymbol{\theta}}(\boldsymbol{z}_i,i)) +  \sqrt{1-\alpha_{i-\frac{1}{2}}^2}\hat{\boldsymbol{\epsilon}}_{\boldsymbol{\theta}}(\boldsymbol{z}_{i}, i).
\label{equ:z_0.5i_solver2}
\end{align}

LA-DPM-Solver-2 is designed by simply replacing $\hat{\boldsymbol{x}}(\boldsymbol{z}_i,\hat{\boldsymbol{\epsilon}}_{\boldsymbol{\theta}}(\boldsymbol{z}_i,i)) $ in (\ref{equ:z_0.5i_solver2}) with an extrapolation, given by 
\begin{align}
    \boldsymbol{z}_{i-1} &= \alpha_{{i-\frac{1}{2}}}\left[(1+\lambda_i)\hat{\boldsymbol{x}}(\boldsymbol{z}_i,\hat{\boldsymbol{\epsilon}}_{\boldsymbol{\theta}}(\boldsymbol{z}_i,i)) - \lambda_i \hat{\boldsymbol{x}}\left(\boldsymbol{z}_{i+\frac{1}{2}},\hat{\boldsymbol{\epsilon}}_{\boldsymbol{\theta}}\left(\boldsymbol{z}_{i+\frac{1}{2}},i+\frac{1}{2}\right)\right)\right] +  \sqrt{1-\alpha_{i-\frac{1}{2}}^2}\hat{\boldsymbol{\epsilon}}_{\boldsymbol{\theta}}(\boldsymbol{z}_{i}, i).
\end{align}
The other quantities in LA-DPM-Solver-2 are computed in the same manner as DPM-Solver-2.

\subsection{LA-DPM-Solver-3}
\label{appendix:LA_Solver_3}

We first present the update expressions for DPM-Solver-3 from \cite{Lu22DPM_Solver}. 

\begin{align}
   \left\{\begin{array}{l} t_{i-\frac{1}{3}} = t_{\lambda}(\frac{\lambda_{t_{i-1}}+2\lambda_{t_i}}{3})  \\
   t_{i-\frac{2}{3}} = t_{\lambda}(\frac{2\lambda_{t_{i-1}}+\lambda_{t_i}}{3})  \\
    \boldsymbol{z}_{i-\frac{1}{3}} =  \frac{\alpha_{{i-\frac{1}{3}}}}{\alpha_{i}}\boldsymbol{z}_{i} - \sigma_{i-\frac{1}{3}}(e^{\frac{h_i}{3}}-1)\hat{\boldsymbol{\epsilon}}_{\boldsymbol{\theta}}(\boldsymbol{z}_{i}, i)  \\
   \boldsymbol{r}_{i-\frac{1}{3}} =  \hat{\boldsymbol{\epsilon}}_{\boldsymbol{\theta}}(\boldsymbol{z}_{i-\frac{1}{3}}, i-\frac{1}{3}) - \hat{\boldsymbol{\epsilon}}_{\boldsymbol{\theta}}(\boldsymbol{z}_{i}, i)   \\
      \boldsymbol{z}_{i-\frac{2}{3}} =  \frac{\alpha_{{i-\frac{2}{3}}}}{\alpha_{i}}\boldsymbol{z}_{i} - \sigma_{i-\frac{2}{3}}(e^{\frac{2h_i}{3}}-1)\hat{\boldsymbol{\epsilon}}_{\boldsymbol{\theta}}(\boldsymbol{z}_{i}, i) - 2\sigma_{i-\frac{2}{3}}\left(\frac{e^{2h_i/3}-1}{(2h_i)/3}-1\right)\boldsymbol{r}_{i-\frac{1}{3}} \\
      \boldsymbol{r}_{i-\frac{2}{3}} =  \hat{\boldsymbol{\epsilon}}_{\boldsymbol{\theta}}(\boldsymbol{z}_{i-\frac{2}{3}}, i-\frac{2}{3}) - \hat{\boldsymbol{\epsilon}}_{\boldsymbol{\theta}}(\boldsymbol{z}_{i}, i)   \\
    \boldsymbol{z}_{i-1} \hspace{-0.8mm}=\hspace{-0.2mm} \frac{\alpha_{i-1}}{\alpha_{i}}\boldsymbol{z}_{i} - \sigma_{i-1}(e^{h_i}-1)\hat{\boldsymbol{\epsilon}}_{\boldsymbol{\theta}}(\boldsymbol{z}_i, i)  - \frac{3\sigma_{i-1}}{2}\left(\frac{e^{h_i}-1}{h_i}-1\right)\boldsymbol{r}_{i-\frac{2}{3}},
     \end{array}\right. \label{equ:solver3}
\end{align}
where $\lambda_t =\log (\alpha_t/\sigma_t)$ is a strictly decreasing function and $t_{\lambda}(\cdot)$ is the reverse function of $\lambda_t$, and $h_i = \lambda_{t_{i-1}} - \lambda_{t_i}$. The two timestep $t_{i-\frac{1}{3}}$ and  $t_{i-\frac{2}{3}}$ are in between $t_i$ and $t_{i-1}$. \textcolor{blue}{It clear from (\ref{equ:solver3}) that the computation of $\boldsymbol{z}_{i-1}$ involves a linear combination of $\hat{\boldsymbol{\epsilon}}_{\boldsymbol{\theta}}(\boldsymbol{z}_i, i)$ and the difference vector $\boldsymbol{r}_{i-\frac{2}{3}} =  \hat{\boldsymbol{\epsilon}}_{\boldsymbol{\theta}}(\boldsymbol{z}_{i-\frac{2}{3}}, i-\frac{2}{3}) - \hat{\boldsymbol{\epsilon}}_{\boldsymbol{\theta}}(\boldsymbol{z}_{i}, i)$ }. 

Next, we study the update expression for $\boldsymbol{z}_{i-\frac{1}{3}}$ in (\ref{equ:solver3}), which can be reformulated as 
\begin{align}
\boldsymbol{z}_{i-\frac{1}{3}} &=  \frac{\alpha_{{i-\frac{1}{3}}}}{\alpha_{i}}\boldsymbol{z}_{i} - \sigma_{i-\frac{1}{3}}(e^{\frac{h_i}{3}}-1)\hat{\boldsymbol{\epsilon}}_{\boldsymbol{\theta}}(\boldsymbol{z}_{i}, i) \nonumber \\
&\textcolor{blue}{\left[\lambda_{t_{i-\frac{1}{3}}} = \frac{\lambda_{t_{i-1}}+2\lambda_{t_i}}{3},\textrm{which is obtained from definition of } t_{i-\frac{1}{3}} \textrm{ in } (\ref{equ:solver3}) \right]} \nonumber \\
&= \frac{\alpha_{{i-\frac{1}{3}}}}{\alpha_{i}}\boldsymbol{z}_{i} - \sigma_{i-\frac{1}{3}}\left(e^{\left(\lambda_{t_{i-\frac{1}{3}}}-\lambda_{t_i} \right) }-1\right)\hat{\boldsymbol{\epsilon}}_{\boldsymbol{\theta}}(\boldsymbol{z}_{i}, i) \nonumber \\  
&\textcolor{blue}{\left[\lambda_{t_{i-\frac{1}{3}}} = \frac{\lambda_{t_{i-1}}+2\lambda_{t_i}}{3}=\log(\alpha_{i-\frac{1}{3}}/\sigma_{i-\frac{1}{3}}),\hspace{2mm} \lambda_{t_i} =\log(\alpha_i/\sigma_i) \right]} \nonumber \\
&= \frac{\alpha_{{i-\frac{1}{3}}}}{\alpha_{i}}\boldsymbol{z}_{i} - \sigma_{i-\frac{1}{3}}\left(\frac{\alpha_{i-\frac{1}{3}}}{\sigma_{i-\frac{1}{3}}}\frac{\sigma_{i}}{\alpha_{i}}  -1\right)\hat{\boldsymbol{\epsilon}}_{\boldsymbol{\theta}}(\boldsymbol{z}_{i}, i) \nonumber \\  
&= \alpha_{{i-\frac{1}{3}}}\underbrace{\left(\frac{\boldsymbol{z}_{i}}{ \alpha_{i}} - \frac{\sigma_i}{\alpha_i} \hat{\boldsymbol{\epsilon}}_{\boldsymbol{\theta}}(\boldsymbol{z}_{i}, i)\right)}_{\hat{\boldsymbol{x}}(\boldsymbol{z}_i,\hat{\boldsymbol{\epsilon}}_{\boldsymbol{\theta}}(\boldsymbol{z}_i,i))} + \sigma_{i-\frac{1}{3}}\hat{\boldsymbol{\epsilon}}_{\boldsymbol{\theta}}(\boldsymbol{z}_{i}, i) \nonumber \\
&\textcolor{blue}{\left[\textrm{For variance preserving process: }\sigma_{i-\frac{1}{3}} =\sqrt{1-\alpha_{i-\frac{1}{3}}^2}\right] } \nonumber \\
&= \alpha_{{i-\frac{1}{3}}}\hat{\boldsymbol{x}}(\boldsymbol{z}_i,\hat{\boldsymbol{\epsilon}}_{\boldsymbol{\theta}}(\boldsymbol{z}_i,i)) +  \sqrt{1-\alpha_{i-\frac{1}{3}}^2}\hat{\boldsymbol{\epsilon}}_{\boldsymbol{\theta}}(\boldsymbol{z}_{i}, i).
\label{equ:z_0.3i_solver3}
\end{align}

To obtain the update expressions of LA-PDM-Solver-3, we modify (\ref{equ:z_0.3i_solver3}) to be 
\begin{align}
\boldsymbol{z}_{i-\frac{1}{3}} =  \alpha_{{i-\frac{1}{3}}}\left[(1+\lambda_i)\hat{\boldsymbol{x}}(\boldsymbol{z}_i,\hat{\boldsymbol{\epsilon}}_{\boldsymbol{\theta}}(\boldsymbol{z}_i,i)) - \lambda_i \hat{\boldsymbol{x}}(\boldsymbol{z}_{i+\frac{1}{3}},\hat{\boldsymbol{\epsilon}}_{\boldsymbol{\theta}}(\boldsymbol{z}_{i+\frac{1}{3}},i+\frac{1}{3}))  \right] +  \sqrt{1-\alpha_{i-\frac{1}{3}}^2}\hat{\boldsymbol{\epsilon}}_{\boldsymbol{\theta}}(\boldsymbol{z}_{i}, i).   
\end{align}
The computation for other quantities in LA-DPM-Solver-3 is the same as in DPM-Solver-3. 

 \subsection{Evaluation of lookahead high-order DPM-Solvers}
\label{appendix:LA_DPM_evaluation}
 \noindent\textbf{Experimental setups}: In this experiment, we took two high-order DPM-solvers from \cite{Lu22DPM_Solver} as two reference methods, which are DPM-Solver-2 and DPM-Solver-3. The two solvers essentially conduct extrapolation on the predicted Gaussian noises to improve the sampling quality. Our objective is to find out if their sampling quality can be further improved by performing additional extrapolation on the estimates of $\boldsymbol{x}$. 

We utilized the same pre-trained model over CIFAR10 for evaluating tAB-DEIS and LA-tAB-DEIS in Subsection~\ref{subsec:DEIS_PNDM} (see Table~\ref{tab:sampling_model}). It is noted that the two high-order solvers in \cite{Lu22DPM_Solver} were designed to work under a small number of timesteps (below 50 in the experiment of \cite{Lu22DPM_Solver}). Therefore, in our experiment, the tested sampling steps are in the range of $[10,40]$. In our improved methods, the strengths of the extrapolations were set to $\lambda_i=0.1$, $i<N$.

\begin{figure}[h!]
\centering
\includegraphics[width=120mm]{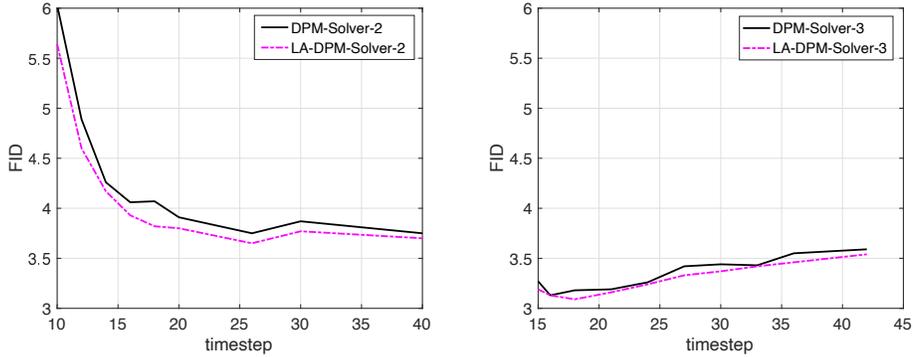}
\vspace*{-0.2cm}
\caption{\footnotesize{Performance of DPM-Solvers and LA-DPM-Solvers for CIFAR10. }}
\label{fig:FID_DPM_Solver}
\vspace{-0.0cm}
\end{figure}

\noindent\textbf{Performance comparison}: Fig.~\ref{fig:FID_DPM_Solver} summarises the FID scores versus timesteps. It is clear that even for high-order DPM-Solvers, the additional extrapolation on estimates of $\boldsymbol{x}$ helps to  achieve lower FID scores. We can also conclude from the figure that DPM-Solver-3 outperforms DPM-Solver-2. This might be because DPM-Solver-3 manages to approximate the integration of the ODE (\ref{equ:ODE}) more accurately than DPM-Solver-2.

We note that Fig.~\ref{fig:FID_DPM_Solver} and Fig.~\ref{fig:DEIS_compare} are based on the same pre-trained model for CIFAR10. By inspection of the FID scores in the two figures, it is clear that tAB-DEIS (order 3) performs better than DPM-Solver-3 for this particular pre-trained model. It is interesting from Fig.~\ref{fig:DEIS_compare} that LA-tAB-DEIS outperforms tAB-DEIS significantly while performance gain of LA-DPM-Solver-3 over DPM-Solver-3 is moderate. The above results demonstrate that the performance gain of our lookahead technique depends on the original sampling method.

\section{Design of LA-S-PNDM}
\label{appendix:LA_S_PNDM}

We summarize the sampling procedure of LA-S-PNDM in Alg.~\ref{alg:LA-S-PNDM}. The only difference between LA-S-PNDM and S-PNDM is the computation of $\boldsymbol{z}_{i-1}$ for $i=N-1,\ldots, 1$. It is seen from Alg.~\ref{alg:LA-S-PNDM} that an additional extrapolation is introduced in terms of the estimates  $\hat{\boldsymbol{x}}_{[i:i+1]}$ and  $\tilde{\boldsymbol{x}}_{[i+1:i+2]}$ of the original data sample $\boldsymbol{x}$. The strengths of the extrapolations are parameterized by $\{\lambda_i\}_{i=1}^{N-1}$. When $\lambda_i=0$ for all $i$, LA-S-PNDM reduces to S-PNDM. 

From Alg.~\ref{alg:LA-S-PNDM}, we observe that the method S-PNDM or LA-S-PNDM  exploits 2nd order polynomial of the estimated Gaussian noises $\{\hat{\boldsymbol{\epsilon}}_{\boldsymbol{\theta}}(\boldsymbol{z}_{i+j}, i+j)\}_{j=0}^1$ in estimation of $\boldsymbol{z}_{i-1}$ at timestep $i$. The polynomial coefficients are computed differently for $i=N$ and $i<N$.

\begin{minipage}{0.63\linewidth}
\begin{algorithm}[H]
\begin{algorithmic}
\caption{Sampling of LA-S-PNDM}
\label{alg:LA-S-PNDM}
\STATE {\small \textbf{Input:}  $\boldsymbol{z}_N\sim \mathcal{N}(\boldsymbol{0}, \boldsymbol{I})$, $\{1>\lambda_i\geq 0\}_{i=1}^{N-1}$ } 
\FOR{$i=N$}
\STATE $(a)\left\{\begin{array}{l}\boldsymbol{z}_{i-1} \hspace{-0.6mm}= \frac{\alpha_{i-1}}{\alpha_i} \left(\boldsymbol{z}_i \hspace{-0.8mm}-\hspace{-0.8mm} \sqrt{1-\alpha_i^2}\hat{\boldsymbol{\epsilon}}_{\boldsymbol{\theta}}(\boldsymbol{z}_i, i) \right)\hspace{-0.1mm}+\hspace{-0.1mm}\sqrt{1-\alpha_{i-1}^2}\hat{\boldsymbol{\epsilon}}_{\boldsymbol{\theta}}(\boldsymbol{z}_i, i) \\
\hat{\boldsymbol{\epsilon}}_{[i-1:i]}=\frac{1}{2}(\hat{\boldsymbol{\epsilon}}_{\boldsymbol{\theta}}(\boldsymbol{z}_i,i)+\hat{\boldsymbol{\epsilon}}_{\boldsymbol{\theta}}(\boldsymbol{z}_{i-1},i-1) ) \\
\hat{\boldsymbol{x}}_{i} = (\boldsymbol{z}_i-\sqrt{1-\alpha_i^2} \hat{\boldsymbol{\epsilon}}_{[i-1:i]})/\alpha_i \\
\boldsymbol{z}_{i-1} \hspace{-0.6mm}= \alpha_{i-1} 
 \hat{\boldsymbol{x}}_{i}   \hspace{-0.1mm}+\hspace{-0.1mm}\sqrt{1-\alpha_{i-1}^2}\hat{\boldsymbol{\epsilon}}_{[i-1:i]} \end{array}\right.$
\ENDFOR
\STATE Denote $\tilde{\boldsymbol{x}}_{[N:N+1]}=\hat{\boldsymbol{x}}_{N} $
\FOR{\small $i=N-1, \ldots, 1$} 
\STATE {\small $(b)\left\{\begin{array}{l}\tilde{\boldsymbol{\epsilon}}_{[i:i+1]}=\frac{1}{2}(3\hat{\boldsymbol{\epsilon}}_{\boldsymbol{\theta}}(\boldsymbol{z}_i,i)-\hat{\boldsymbol{\epsilon}}_{\boldsymbol{\theta}}(\boldsymbol{z}_{i+1},i+1) ) \\
\hat{\boldsymbol{x}}_{[i:i+1]} = (\boldsymbol{z}_i-\sqrt{1-\alpha_i^2} \tilde{\boldsymbol{\epsilon}}_{[i:i+1]})/\alpha_i \\
\boldsymbol{z}_{i-1} \hspace{-0.6mm}= \alpha_{i-1} \textcolor{blue}{\left((1+\lambda_i)\hat{\boldsymbol{x}}_{[i:i+1]} - \lambda_i\tilde{\boldsymbol{x}}_{[i+1:i+2]} \right)}\hspace{-0.1mm}+\hspace{-0.1mm}\sqrt{1-\alpha_{i-1}^2}\tilde{\boldsymbol{\epsilon}}_{[i:i+1]} \\
\tilde{\boldsymbol{x}}[i:i+1]=(1+\lambda_i)\hat{\boldsymbol{x}}_{[i:i+1]} - \lambda_i\tilde{\boldsymbol{x}}_{[i+1:i+2]}\end{array}
\right.$ }

\ENDFOR        
\STATE {\small \textbf{output:} $\boldsymbol{z}_0$ } \\
\noindent\rule{\textwidth}{0.5pt}
\vspace{-2.5mm}
\item * The update for $\boldsymbol{z}_{N-1}$ in $(a)$ is referred to as pseudo improved Euler step in \cite{Liu22PNDM}. 
\item * The update for $\boldsymbol{z}_{i-1}$ in $(b)$ is referred to as pseudo linear multi step in \cite{Liu22PNDM}. 
\item * LA-S-PNDM reduces to S-PNDM when $\{\lambda_i=0\}_{i=N-1}^1$.
\vspace{1mm}
\end{algorithmic}
\end{algorithm}
\end{minipage}

\section{Tested Pre-trained Models in Experiments }
\label{appendix:sampling_model}

 \begin{table}[h!]
\caption{\small sampling methods and the corresponding pre-trained models \vspace{0mm} } 
\label{tab:sampling_model}
\centering
\begin{tabular}{|c|c|}
\hline
\small{sampling methods} & {model name} \\
\hline
\small{$\begin{array}{l}\textrm{Fig.~\ref{fig:DEIS_compare} for tAB-DEIS and LA-tAB-DEIS}
\\
\textrm{Fig.~\ref{fig:FID_DPM_Solver} for DPM-Solvers and LA-DPM-Solvers}
\end{array}$}  &\small{$\begin{array}{c}\textrm{cifar10}\_\textrm{ddpmpp}\_\textrm{deep}\_\textrm{continuous/checkpoint}\_\textrm{8.pth } \\ \textrm{(from https://github.com/yang-song/score}\_\textrm{sde)}\end{array}$ } \\
\hline 
\small{Fig.~\ref{fig:SPNDM} for S-PNDM and LA-S-PNDM} & \small{$\begin{array}{l}1. \textrm{ddim}\_\textrm{cifar10.ckpt} \\ 
2. \textrm{ddim}\_\textrm{celeba.ckpt} \\ 
3. \textrm{ddim}\_\textrm{lsun}\_\textrm{bedroom.ckpt}\\ 
4. \textrm{ddim}\_\textrm{lsun}\_\textrm{church.ckpt}  \\
\textrm{(from \url{https://github.com/luping-liu/PNDM})}\end{array}$ } \\
\hline 
\end{tabular}
\end{table}

\begin{figure}[h!]
\centering
\includegraphics[width=160mm]{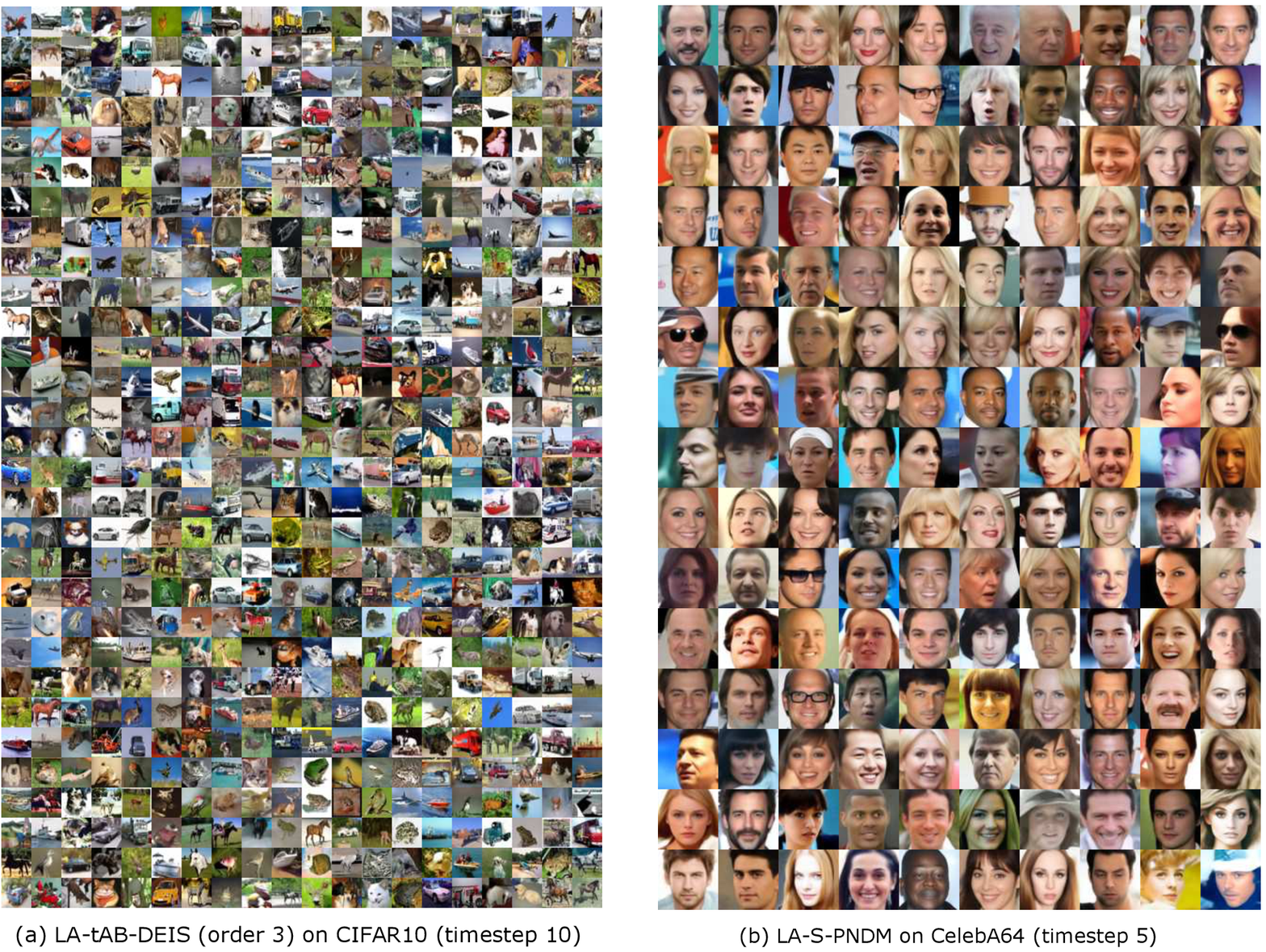}
\vspace*{-0.2cm}
\caption{\small{Generated images with LA-tAB-DEIS and LA-S-PNDM}}
\label{fig:FID_img_deis_pndm_cifar10_celeba}
\vspace{-0.0cm}
\end{figure}

\begin{figure}[h!]
\centering
\includegraphics[width=160mm]{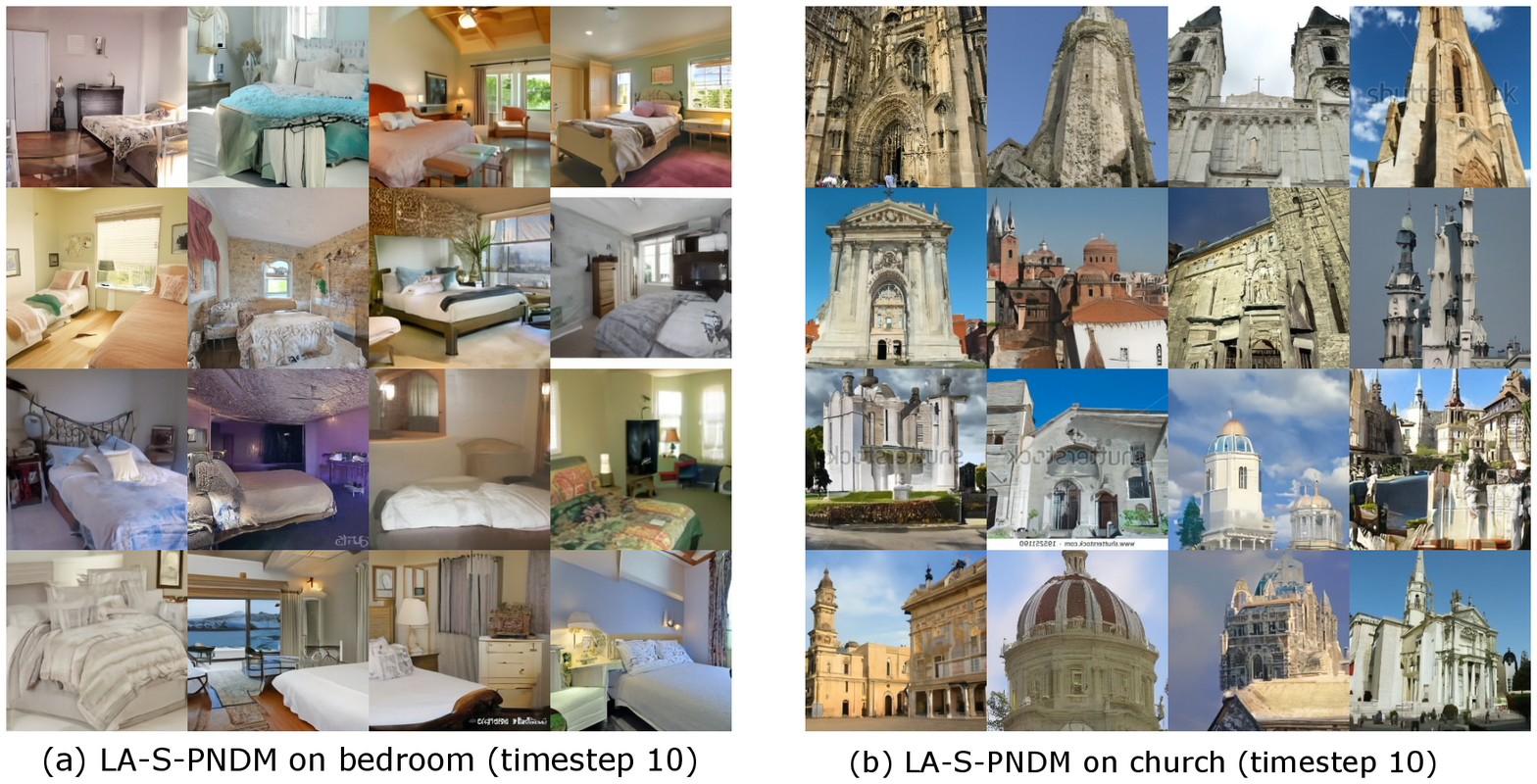}
\vspace*{-0.2cm}
\caption{\small{Generated images with LA-S-PNDM}}
\label{fig:FID_img_pndm_bedroom_church}
\vspace{-0.0cm}
\end{figure}

\end{document}